%% file: iclr2024_conference.tex
\pdfoutput=1

\documentclass[11pt]{article}

\usepackage[dvipsnames]{xcolor}

\usepackage[preprint]{acl}

\usepackage{times}
\usepackage{latexsym}

\usepackage[T1]{fontenc}

\usepackage[utf8]{inputenc}

\usepackage{microtype}

\usepackage{inconsolata}

\input{math_commands.tex}

\usepackage{hyperref}

\usepackage{longtable}
\usepackage{bbm}
\usepackage[shortlabels]{enumitem}
\usepackage{algorithm}
\usepackage{algorithmicx} 
\usepackage{algpseudocode} 

\usepackage{graphicx}
\usepackage{multirow}
\usepackage{booktabs}
\usepackage{colortbl}
\usepackage{mdframed}
\usepackage{caption}
\usepackage{subfigure}
\usepackage{wrapfig}
\usepackage{listings}
\usepackage{lipsum}  
\usepackage{longtable}
\usepackage{bbding}

\colorlet{punct}{red!60!black}
\definecolor{background}{HTML}{EEEEEE}
\definecolor{delim}{RGB}{20,105,176}
\colorlet{numb}{magenta!60!black}

\lstdefinelanguage{json}{
    basicstyle=\scriptsize\ttfamily,
    numbers=left,
    numberstyle=\scriptsize,
    stepnumber=1,
    numbersep=8pt,
    xleftmargin=16pt,
    showstringspaces=false,
    breaklines=true,
    frame=lines,
    backgroundcolor=\color{background},
    literate=
     *{0}{{{\color{numb}0}}}{1}
      {1}{{{\color{numb}1}}}{1}
      {2}{{{\color{numb}2}}}{1}
      {3}{{{\color{numb}3}}}{1}
      {4}{{{\color{numb}4}}}{1}
      {5}{{{\color{numb}5}}}{1}
      {6}{{{\color{numb}6}}}{1}
      {7}{{{\color{numb}7}}}{1}
      {8}{{{\color{numb}8}}}{1}
      {9}{{{\color{numb}9}}}{1}
      {:}{{{\color{punct}{:}}}}{1}
      {,}{{{\color{punct}{,}}}}{1}
      {\{}{{{\color{delim}{\{}}}}{1}
      {\}}{{{\color{delim}{\}}}}}{1}
      {[}{{{\color{delim}{[}}}}{1}
      {]}{{{\color{delim}{]}}}}{1},
}

\newcommand\blfootnote[1]{%
  \begingroup
  \renewcommand\thefootnote{}\footnote{#1}%
  \addtocounter{footnote}{-1}%
  \endgroup
}

\definecolor{color1}{HTML}{f9e086}
\definecolor{color2}{HTML}{f7c1ad}
\definecolor{color3}{HTML}{cbe6e2}

\title{Prompt Engineering a Prompt Engineer}

\author{Qinyuan Ye$^{1\dagger}$ \ \ Maxamed Axmed$^{2}$\ \ Reid Pryzant$^{2}$\ \ Fereshte Khani$^{2}$ \\
$^{1}$University of Southern California \ \ 
$^{2}$Microsoft\\
\texttt{qinyuany@usc.edu} \ \ \texttt{fkhani@microsoft.com}
}

\begin{document}

\maketitle

\begin{abstract}
Prompt engineering is a challenging yet crucial task for optimizing the performance of large language models on customized tasks. 
It requires complex reasoning to examine the model's errors, hypothesize what is missing or misleading in the current prompt, and communicate the task with clarity. 
While recent works indicate that large language models can be meta-prompted to perform automatic prompt engineering, we argue that their potential is limited due to insufficient guidance for complex reasoning in the meta-prompt.
We fill this gap by infusing into the meta-prompt three key components: detailed descriptions, context specification, and a step-by-step reasoning template.
The resulting method, named PE2, exhibits remarkable versatility across diverse language tasks. It finds prompts that outperform ``let's think step by step'' by 6.3\% on MultiArith and 3.1\% on GSM8K, and outperforms competitive baselines on counterfactual tasks by 6.9\%.
Further, we show that PE2 can make targeted and highly specific prompt edits, rectify erroneous prompts, and induce multi-step plans for complex tasks.\blfootnote{\hspace{-0.13cm}$^\dagger$Work done while interning at Microsoft.}
\end{abstract}

\section{Introduction}

\begin{figure}[t]
    \centering
    \includegraphics[width=0.48\textwidth]{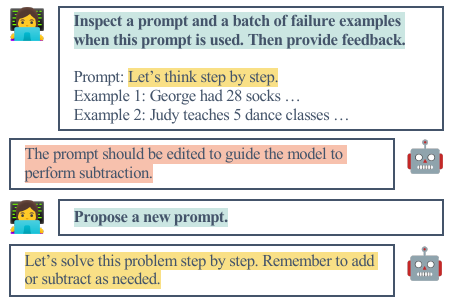}
    \caption{LLM-powered automatic prompt engineering methods typically use a \colorbox{color3}{meta-prompt} that guides an LLM to inspect the \colorbox{color1}{current prompt}, provide \colorbox{color2}{feedback} (sometimes refered to as textual ``gradients'') and then generate an \colorbox{color1}{updated prompt}. In this paper, we design and investigate \colorbox{color3}{meta-prompt} variants to guide LLMs to perform automatic prompt engineering more effectively.  
    }
    \label{fig:teaser}
\end{figure}

Large language models (LLMs) are powerful tools for many natural language processing tasks, when provided with the right prompts.\footnote{In this paper, we focus on textual prompts of task description (\textit{e.g.}, ``Translate English to French'') or instruction (\textit{e.g.}, ``Let's think step by step'').} 
However, LLMs are also notoriously sensitive to prompt design \citep{jiang-etal-2020-know,Zhao2021CalibrateBU,reynolds2021prompt,lu-etal-2022-fantastically}, and finding the right prompts often requires extensive manual efforts referred to as ``prompt engineering.''
Non-AI experts, in particular, may struggle to effectively communicate the task of interest to LLMs, resulting in prompt engineering being performed opportunistically rather than systematically \cite{10.1145/3544548.3581388}.
Adding to this challenge, once a high-quality prompt is found and deployed into production, unforeseen edge cases can arise, necessitating more rounds of manual efforts.
All these challenges give rise to an emerging research field of \textit{automatic} prompt engineering. Within this field, a notable line of methods involves leveraging the capabilities of LLMs themselves \citep{zhou2023large, pryzant-etal-2023-automatic}. 
This entails meta-prompting LLMs with instructions such as ``inspect the current prompt and a batch of examples, provide feedback, then propose a new prompt.'' (See Figure~\ref{fig:teaser})

\begin{figure*}[t]
    \centering
\includegraphics[width=\textwidth]{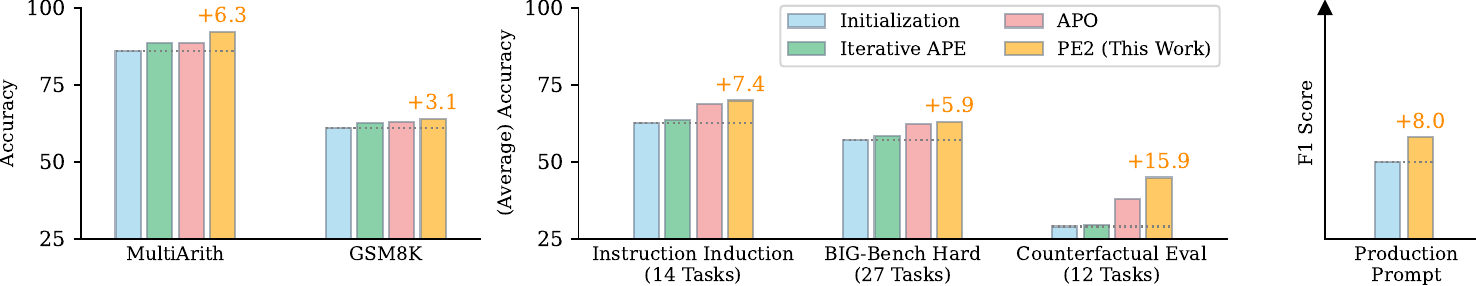}\vspace{-0.2cm}
\caption{\textbf{Results Overview.} Our method PE2 consistently brings improvements over the prompt initialization (marked with {\color{YellowOrange} orange text}). It matches or outperforms prompt optimization baselines Iterative APE \citep{zhou2023large} and APO \citep{pryzant-etal-2023-automatic} across a wide range of language tasks, with most significant performance gain observed on Conterfactual Eval \cite{wu2023reasoning}. See detailed performance breakdown in Fig.~\ref{fig:ii}-\ref{fig:bbh}.
}
\label{fig:results_overview}
\end{figure*}

While these methods achieve impressive performance, a subsequent question arises: What makes a good meta-prompt for automatic prompt engineering? To answer this question, we connect two key observations: (1) Prompt engineering, at its core, is a language generation task that requires complex reasoning: it involves closely examining the model's errors, hypothesizing what is missing or misleading in the current prompt, and communicating the task more clearly to LLMs. (2) Complex reasoning capabilities in LLMs can be elicited by prompting the model to ``think step by step'' \citep{wei2022cot,kojima2022large} and can be further improved by instructing them to reflect on their outputs \citep{madaan2023self, chen2023teaching}.

Bridging these two observations, in this work, we prompt engineer a prompt engineer---we aim to construct a meta-prompt that guide LLMs to perform automatic prompt engineering more effectively.
We argue that prior works do not provide sufficient guidance in the meta-prompt, thereby limiting the potential of LLMs for automatic prompt engineering. 
To address this, we introduce new meta-prompt components such as detailed two-step task descriptions, context specification, and a step-by-step reasoning template, to better equip LLMs throughout the process  (\S\ref{sec:pe2}; Fig.~\ref{fig:intro}).

The resulting method, named PE2, achieves strong empirical performance (\S\ref{ssec:empirical-results}). When using \texttt{text-davinci-003} as the task model, the prompts produced by PE2 surpass the zero-shot chain-of-thought prompt ``let's think step by step" \citep{kojima2022large} by 6.3\% on MultiArith and 3.1\% on GSM8K. Moreover, PE2 matches or outperforms two automatic prompt engineering baselines, Iterative APE \citep{zhou2023large} and APO \citep{pryzant-etal-2023-automatic} in multiple settings (Fig.~\ref{fig:results_overview}). PE2 is most effective on counterfactual tasks \citep{wu2023reasoning}, where the automatic prompt engineer is anticipated to reason about non-standard situations (\textit{e.g.}, do addition in base-8 instead of base-10) and explain such situation to the task model through the prompt. Beyond academic datasets, we show that PE2 can improve an expert-written production prompt consisting of over 5,000 tokens, resulting in an 8.0\% increase in F1 score.

We further provide a detailed analysis on the behaviors, advantages, and limitations of PE2.
Upon examining the prompt edit history (\S\ref{ssec:pe2_behavior}), we find that PE2 consistently offers meaningful prompt edits (Table~\ref{table:case_study}). It is able to amend erroneous or incomplete prompts and enrich the prompts with additional details, which leads to improved final performance. 
It is also able to devise multi-step plans for complex tasks. For example, in the task of movie recommendation, PE2 makes the plan to ``consider factors such as genre, plot and style'' in the prompt.
Interestingly, when uninformed about performing addition in base-8, PE2 formulates partially-correct arithmetic rules from examples by itself: ``If both numbers are less than 50, add 2 to the sum. If either number is 50 or greater, add 22 to the sum.\footnote{Both the base-8 addition rules and the model-induced rules hold true for examples like 75+7=104 and 5+6=13.}''
This demonstrates PE2's remarkable ability to reason and adapt in non-standard situations, while also raises concerns of ``shortcut learning'' in prompt optimization. 

\section{Background}
In this section, we provide a formal formulation of the prompt engineering problem (\S\ref{ssec:pe}), and describe a general framework of automatic prompt engineering using LLMs and meta-prompts (\S\ref{ssec:automatic}). 
Building on this foundation, we introduce new meta-prompt components used in PE2 in \S\ref{sec:pe2}.

\subsection{Prompt Engineering}
\label{ssec:pe}

The goal of prompt engineering is to find the textual prompt $p^*$ that achieves the best performance on a given dataset $D$ when using a given LLM $\mathcal{M}_{task}$ as the task model. 
More specifically, we assume all datasets can be formatted as textual input-output pairs, \textit{i.e.}, $D=\{(x,y)\}$. 
We are given a training set $D_{train}$ for optimizing the prompt, $D_{dev}$ for validation, and $D_{test}$ for final evaluation. 
Following the notations in \citet{zhou2023large}, the prompt engineering problem can be described as:
\begin{equation}
    p^* = \argmax_{p} \sum_{(x,y)\in D_{dev}} f(\mathcal{M}_{task}(x; p), y)\label{eq:pe}
\end{equation}
where $\mathcal{M}_{task}(x; p)$ is the output generated by the task model when conditioning on the prompt $p$, and $f$ is a per-example evaluation function. For example, if the evaluation metric is exact match, $f(\mathcal{M}_{task}(x; p), y) = \mathbbm{1}[\mathcal{M}_{task}(x; p) = y]$.

\subsection{Automatic Prompt Engineering with LLMs}
\label{ssec:automatic}

To alleviate the intensive efforts of manual prompt engineering, recent works explore automating this process by meta-prompting LLMs to paraphrase the prompt \citep{zhou2023large} or refine the prompt by inspecting failure examples \citep{pryzant-etal-2023-automatic}. In the following, we describe a framework that encapsulates these prior works and is employed in our development of PE2 in later sections.
It has three parts: prompt initialization, new prompt proposal, and the search procedure.

\paragraph{Prompt Initialization.} To start the prompt engineering process, a set of initial prompts $P^{(0)}$ is needed.
We consider two initialization methods:
\textbf{(1) Manual initialization} is applicable for tasks that has pre-existing prompts written by humans experts. For example, ``Let's think step by step'' is effective on mathematical reasoning tasks and can be used as the initialization for prompt optimization.
In \textbf{(2) Induction Initialization}, we follow \citet{zhou2023large} by using a batch of examples $\{(x,y)\}$ from $D_{train}$ and a prompt $p^{init}$ (``Here are the input-output pairs. What is the instruction?'') to generate a set of initial prompts $P^{(0)}$.

\paragraph{New Prompt Proposal.} 
Given a set of initial prompts, the automatic prompt engineer will continuously propose new and potentially better prompts. At timestamp $t$, the prompt engineer is given a prompt $p^{(t)}$ and expected to write a new prompt $p^{(t+1)}$. Optionally, a batch of examples $B=\{(x, y, y')\}$ may be inspected in the new prompt proposal process. Here $y'=\mathcal{M}_{task}(x; p)$ represents output generated by the task model and $y$ represents the ground-truth label. We use $p^{meta}$ to denote a meta-prompt that is used to instruct the prompt proposal model 
$\mathcal{M}_{proposal}$ to propose new prompts. Therefore, 
\begin{equation}
    p^{(t+1)} = \mathcal{M}_{proposal}(p^{(t)}, B; p^{meta})
\label{eq:optim}
\end{equation}
Constructing a better meta-prompt $p^{meta}$ to improve the quality of the proposed prompt $p^{(t+1)}$ is the main focus of this study. We will describe in more details in \S\ref{sec:pe2}.

\paragraph{Search Procedure.} 
As LLMs are sensitive to trivial prompt variations, it is possible that the newly proposed prompt $p^{(t+1)}$ under-performs the original prompt $p^{(t)}$. Therefore, automatic prompt engineering is typically combined with a back-tracking enabled search procedure.
At timestamp $t$, we select $n$ best-performing prompts from \textit{all} prompt candidates obtained in previous timestamps (\textit{i.e.}, $P^{(0)}\cup P^{(1)} \cup ... \cup P^{(t)}$). 
For \textit{each} of these $n$ prompts, we sample $m$ different batches $B$ of model errors, and run the meta-prompt in Eq.~\ref{eq:optim} to produce $m$ new prompts.
This results in $m\times n$ new prompts, which we denote as $P^{(t+1)}$ collectively and are used at the next timestamp $t+1$.
The search algorithm is described more formally in Algorithm \ref{algo:search}.

\section{Prompt Engineering a Prompt Engineer}
\label{sec:pe2}
\begin{figure*}[t]
    \centering
    \includegraphics[width=\textwidth]{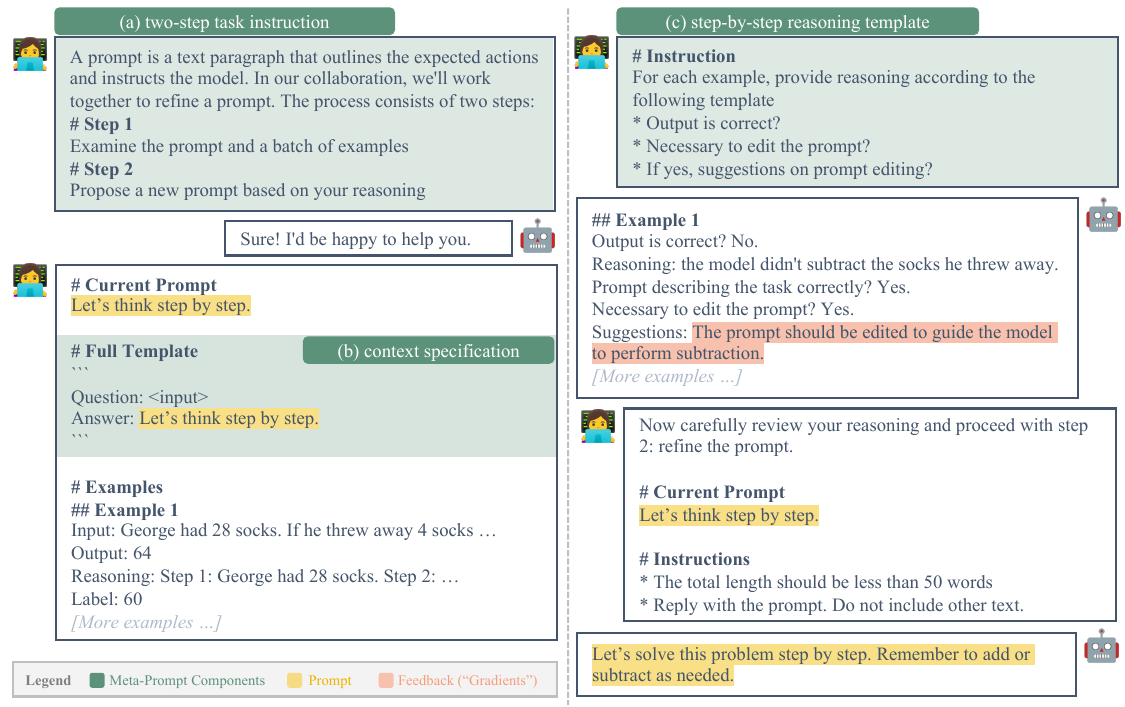}
    \caption{An redacted example to illustrate the meta-prompt components introduced in \S\ref{sec:pe2}. 
    }
    \label{fig:intro}
\end{figure*}

Much like how the prompt plays an important role for the end task performance, the meta-prompt $p^{meta}$ introduced in Eq.~\ref{eq:optim} plays an important role in the quality of newly proposed prompts, and the overall quality of automatic prompt engineering. 
In this work, we focus on prompt engineering the meta-prompt $p^{meta}$---we develop meta-prompt components that can potentially help improve LLMs' prompt engineering quality.

\begin{table}[]
\scalebox{0.75}{
\begin{tabular}{@{}l|ccc@{}}
\toprule                                                             & \textbf{Iter. APE} & \textbf{APO}        & \textbf{PE2}                                                              \\\midrule
Inspect Model Failures                                              & \raisebox{-.5ex}{\XSolidBrush}        & \raisebox{-.5ex}{\Checkmark}        & \raisebox{-.5ex}{\Checkmark}                                                              \\
(a) Task Description                                                    &           &            &                                                                  \\
\quad - Length                                                              & Short     & Short      & Long                                                             \\
\quad - \# Steps                                                              & One-step  & Two-step   & Two-step                                                         \\
\quad - Position & Beginning & On-the-fly & Both \\
(b) Context Specification                                               & \raisebox{-.5ex}{\XSolidBrush}         & \raisebox{-.5ex}{\XSolidBrush}          & \raisebox{-.5ex}{\Checkmark}                                                              \\
(c) Step-by-step Template                                               & \raisebox{-.5ex}{\XSolidBrush}         & \raisebox{-.5ex}{\XSolidBrush}          & \raisebox{-.5ex}{\Checkmark}  \\\bottomrule
\end{tabular}
}
\caption{Comparison of Meta-prompt Components Used in Baseline Methods and PE2.}\label{tab:summary}
\end{table}

In the following, we reflect on the limitations of prior works and subsequently introduce three meta-prompt components targeting these limitations.
We visualize these components in Fig.~\ref{fig:intro} and provide a summary in Table~\ref{tab:summary}. We name our method using these three components as \textbf{PE2}, a prompt engineered prompt engineer.

\paragraph{(a) Two-step Task Description.} In APO \cite{pryzant-etal-2023-automatic} the task of prompt engineering is decomposed into two steps (Fig.~\ref{fig:teaser}): In step 1, the model is expected to inspect the current prompt and a batch. In step 2, the model is expected to generate an improved prompt.

However, in APO, each step is explained \textit{briefly} and \textit{on the fly}. In contrast, we consider clarifying the steps and expectations \textit{upfront} in the meta-prompt, and also guiding the model with specific steps \textit{on-the-fly}.

\paragraph{(b) Context Specification.} In practice, how the prompt and the input text are formatted together is flexible.
It may appear \textit{before} the input text to describe the task, \textit{e.g.}, ``Translate English to French.'' It may appear \textit{after} the input text, \textit{e.g.}, ``Let's think step by step'' to elicit reasoning capabilities.
Recognizing these varying contexts, we explicitly specify the layout of the prompt and the input. 

\paragraph{(c) Step-by-step Reasoning Template.} To encourage the model to examine \textit{each} example in the batch $B$ closely and reflect on the limitations in the current prompt, we guide the prompt proposal model $\mathcal{M}_{proposal}$ with a list of questions. For example:
Is the prompt correctly describing the task? 
Is it necessary to edit the prompt?
If yes, provide actionable suggestions on prompt editing.

\paragraph{Other Meta-prompt Components We Tried.} Inspired by optimization concepts such as batch size, step size and momentum, we considered adding their verbalized counterparts to the meta-prompt and investigate their effects. We also considered adding a prompt engineering tutorial in the meta-prompt to help the LLM better understand the task. Our observations on these components are mixed. We report these results in Appendix~\ref{app:meta-prompt-optim}.

\section{Experiment Setting}
\subsection{Tasks} 
\label{ssec:datasets}
We use the following five groups of tasks to evaluate the effectiveness of PE2. Details of the used datasets (\textit{e.g.}, dataset sizes, train-test splits, license) are deferred in Appendix~\ref{app:data}.

\paragraph{(1) Math Reasoning.} We use MultiArith \citep{roy-roth-2015-solving} and GSM8K \citep{cobbe2021training}, which contain grade school math problems that requires multiple steps of arithmetic operations.

\paragraph{(2) Instruction Induction.} Instruction Induction \citep{honovich-etal-2023-instruction} is a benchmark for inferring the underlying instruction from few-shot examples. We use 14 selected tasks that cover a wide range of use cases, \textit{e.g.}, ``Formality'' is a task that aims at rephrasing a sentence to be more formal.

\paragraph{(3) BIG-bench Hard.} BIG-bench Hard \cite{suzgun-etal-2023-challenging} is a collection of 23 tasks (27 subtasks) that are challenging to LLMs but the performance may be improved with advanced prompting techniques \cite{chain_of_thought}. Some of BIG-bench Hard tasks are closely related to real-world applications (\textit{e.g.}, movie recommendation).

\paragraph{(4) Counterfactual Evaluation.} We use the arithmetic, chess, and syntax tasks and their counterfactual variants introduced in \citet{wu2023reasoning}. For arithmetic, the original task is base-10 addition, and the counterfactual tasks are base-8/9/11/16 addition. For chess, the starting positions for knights and bishops are swapped in the counterfactual task.
We use this set of tasks to investigate whether PE2 can reason about non-standard situations and communicate them to the task model.

\paragraph{(5) Production Prompt.} Lastly, we optimize an internal production prompt for a hierarchical, multi-label classification task. The task is to classify a user query into domains, intents and slots, and then output a nested dictionary as the result. The initialization prompt is carefully designed by experienced engineers and has more than 5,000 tokens.

\subsection{Compared Methods}
\label{sec:exp-details}
We compare PE2 with the following automatic prompt engineering methods.
\textbf{(a) APE} \citep{zhou2023large}: The base version of APE is an initialization-only method and does not involve new prompt proposal steps. It uses an initialization prompt $p^{init}$ to generate multiple prompt candidates from a few examples, and select the best one among them based on $D_{dev}$ performance. 
\textbf{(b) Iterative APE} \citep{zhou2023large}: After initialization, $p^{meta}$ instructs the model to produce a paraphrase of $p^{(t)}$ and use it as $p^{(t+1)}$. \textbf{(c) APO} \citep{pryzant-etal-2023-automatic}: $p^{meta}$ contains short instructions on inspecting the batch $B$, generating textual ``gradients'' (feedback), and producing a new prompt $p^{(t+1)}$. We include the $p^{init}$ and $p^{meta}$ used in these baseline methods in Appendix~\ref{app:meta-prompt}.

\subsection{Expeirment Details}\label{ssec:budget}
\paragraph{LLMs.} By \textit{default}, we use \texttt{gpt-4} \citep{OpenAI2023GPT4TR} as prompt proposal model $\mathcal{M}_{proposal}$ and use \texttt{text-davinci-003} \citep{ouyang2022training} as the task model $\mathcal{M}_{task}$ performing the underlying task. Experiments on BIG-bench Hard are conducted at a later stage, and we use \texttt{gpt-4-turbo} and \texttt{gpt-3.5-turbo-instruct} to save costs and demonstrate the compatibility of our methods. 
To ensure fair comparison, we always use the same set of LLMs ($\mathcal{M}_{proposal}$ and $\mathcal{M}_{task}$) when evaluating PE2 against other prompt optimization methods.

\paragraph{Prompt Initialization.} For Math Reasoning and BIG-bench Hard tasks, we use ``Let's think step by step.'' \citep{kojima2022large} as the initialization prompt, which can elicit multi-step reasoning in LLMs to perform these tasks. For Instruction Induction, we follow the setting in prior works \citep{zhou2023large} and use induction initialization. For Counterfactual Eval, we experiment with both. For the production task, we use the prompt written by an experienced engineer.

\paragraph{Search Budget.} We use the same search budget for all prompt optimization methods.
For experiments using induction initialization, 30 prompts are generated by $p^{init}$ and form the initial candidate set $P^{(0)}$.
Due to budget constraints, the number of optimization steps $T$ is set to be $3$. At each timestamp, we select $n=4$ best-performing prompts, and propose $m=4$ prompts from each of them.

\vspace{0.2cm}
\noindent We defer other experiment details in Appendix~\ref{app:exp_details}.

\section{Results and Analysis}

\subsection{Main Results}
\label{ssec:empirical-results}

\paragraph{Improved baselines with more recent LLMs.}
In Zero-shot CoT \citep{kojima2022large} and APE \citep{zhou2023large}, the results were obtained with a earlier \texttt{text-davinci-002} model. We first rerun the prompts in these two works with \texttt{text-davinci-003}, an upgraded model. In Table \ref{tab:math_results}, we observe a significant performance boost by using \texttt{text-davinci-003}, suggesting that it is more capable of solving math reasoning problems with Zero-shot CoT. Moreover, the gaps between the two prompts are narrowed (MultiArith: $3.3\%\rightarrow1.0\%$, GSM8K: $2.3\%\rightarrow0.6\%$), indicating \texttt{text-davinci-003} has a reduced sensitivity to prompt paraphrasing. 
Given this, methods that rely on simple paraphrasing, such as Iterative APE, may not improve the final accuracy as effectively.
More specific and targeted edits are necessary to improve the performance.

\paragraph{PE2 outperforms Iterative APE and APO on various tasks.} 
PE2 is able to find a prompt that achieves $92.3\%$ accuracy on MultiArith ($+6.3\%$ compared to Zero-shot CoT) and $64.0\%$ on GSM8K ($+3.1\%$). 
Additionally, we demonstrate the wide applicablity of PE2 on a wide range of language tasks. 
In Fig.~\ref{fig:results_overview} we summarize the results and show that PE2 outperforms Iterative APE \citep{zhou2023large} and APO \citep{pryzant-etal-2023-automatic} in multiple cases. 
Most notably, when induction initialization is used, PE2 outperforms APO on 11 out of 12 counterfactual tasks (Fig.~\ref{fig:results-cf}), exhibiting a 6.9\% average increase in accuracy. This highlights PE2's capability in reasoning about contradictions and unconventional situations. We defer experiment details and performance breakdown for these benchmarks in Appendix~\ref{app:additional_results}.

\input{tables/math_results}

\input{tables/mistral_results}

\input{tables/math_prompts}
\begin{figure}[t]
    \centering
    \includegraphics[width=0.49\textwidth, clip, trim=0 0 0 1.2em]{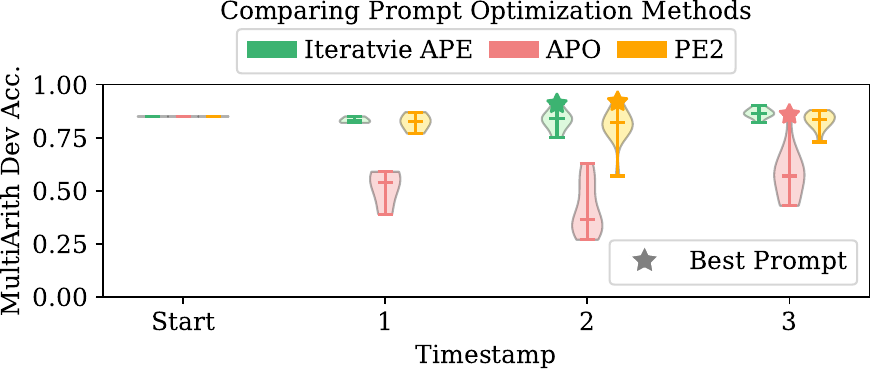}
    \vspace{-0.5cm}
    \caption{Prompt optimization dynamics of Iterative APE, APO and PE2 on MultiArith. The violins represent the quality distributions of newly proposed prompts at each optimization step. PE2 has a better balance between exploration and stability.
    }
    \label{fig:dynamics}
\end{figure}

\paragraph{PE2 generates targeted prompt edits and high-quality prompts.} In Fig.~\ref{fig:dynamics} we plot the quality of prompt proposals over the course of prompt optimization. We observe very distinct patterns for the three prompt optimization methods: 
Iterative APE is based on paraphrasing, so the newly generated prompts have smaller variance. APO makes drastically large prompt edits and thus the performance drops in the first step. Among the three methods, PE2 has a better balance between exploration and stability.
In Table \ref{tab:math_prompts}, we list the optimal prompts found by these methods. Both APO and PE2 are able to provide instructions on ``considering all parts / details''. In addition, PE2 is designed to inspect the batch closely, enabling it make very specific prompt edits such as ``remember to add or subtract as needed''.

\paragraph{PE2 is widely applicable to various LLMs, including open-weight models.}
Our previous experiment setting employs different combinations of \textit{closed-source} models as the task model $\mathcal{T}_{task}$ and the prompt proposal model $\mathcal{T}_{proposal}$ (\S\ref{ssec:budget}).
To further demonstrate PE2's model-agnostic nature, we introduce \texttt{Mistral-7B-Instruct-v0.2}, a recent open-weight model, as the task model, and uses \texttt{gpt-4-turbo} as the prompt proposal model. We report the results in Table~\ref{table:mistral_results} and the final optimized prompts in Table~\ref{table:mistral-prompts}. Consistent with our experiments with closed-source models, results in Table~\ref{table:mistral_results} demonstrate that PE2 performs competitively with or surpasses other automated prompt engineering methods.
As recent research suggests, LLMs still exhibit surprising sensitivity to prompt design and formatting \cite{sclar2024quantifying, mizrahi2023state}, highlighting the importance of investigating why certain prompts succeed or fail. We hope PE2 empowers researchers to discover effective and ineffective prompts, which lay empirical foundations for future exploration into prompt sensitivity with open-weight models.

\subsection{Ablation Study}
\label{ssec:ablation}

To demonstrate the effectiveness of the three new meta-prompt components
introduced in PE2, we run ablation experiments by removing these components during prompt optimization on MultiArith and GSM8K. In these experiments, we make sure that the meta-prompt still contains sufficient information about the task of prompt engineering.
From the results in Table~\ref{tab:ablation}, we show that these three components contribute significantly to the final accuracy. 
In Fig.~\ref{fig:ablation}, we visualize the optimization dynamics of these ablation experiments. 
We find that the exclusion of any one of these components results in a higher variance in the quality distribution of newly-proposed prompts. 
Moreover, without these components, the proposal model more frequently suggests low-quality prompts.

We also conduct an ablation study on back-tracking (\textit{i.e.}, at timestamp $t$, select top-performing prompts from $\cup_{i=0}^t P^{(i)}$ versus only $P^{(t)}$) and hard negative sampling (\textit{i.e.}, the batch $B$ is sampled from the model's errors, versus being randomly sampled from $D_{train}$). 
Since both techniques show slightly positive effects on PE2's performance, they are retained in the final version of PE2. 

We encourage readers to refer to \S\ref{app:meta-prompt-optim} for additional meta-prompt components that we explored during PE2's development, such as verbalized ``momentum'', ``step size'', and a tutorial on prompt engineering. Although these elements were not included in the final version of PE2, we document them to encourage further exploration as more capable language models emerge in the future.

\subsection{Case Study}
\label{ssec:pe2_behavior}

\paragraph{PE2 amends erroneous or incomplete instructions, and devises multi-step plans for complex tasks.} 
Tables~\ref{table:case_study} and~\ref{table:case_study2} present notable prompt edits made by PE2.
In the task of finding rhyming words (\textit{e.g.} ``car'' rhymes with ``bar''), induction initialization mistakenly suggests the task is about changing the first letter of a word.
PE2 successfully correct this after one optimization step. 
In the task of movie recommendation, PE2 is able to decompose the complex task into concrete criteria, such as genre, plot and actor, when determining movie similarities. In date understanding, PE2 identifies the crucial step of referencing information about ``today''.
These examples demonstrate PE2's ability to learn by summarizing key steps from failures and incorporating them into improved prompts, aligning with recent work \cite{zhang2024context}.

\input{tables/ablation}
\input{tables/successful_edits}

\paragraph{Limitations in meta-prompt following and hallucination.} Despite the successes made by PE2, we note several factors that's limiting its performance when applied to challenging counterfactual tasks. We provide representative cases in Table~\ref{table:limitation}. Occasionally, PE2 refuses to propose a new prompt and insists that ``the prompt is correct, but the label is incorrect,'' even when we explicitly state "the labels are absolutely correct" within the meta-prompt.
In another example, we attempt to guide it with hints (\textit{e.g.}, suggesting a different numerical base), but this can unfortunately lead PE2 to generate incorrect solutions (\textit{e.g.}, base-80) and even create justifications for these imagined solutions.
These observations highlight the importance of improving LLMs' abilities to follow (meta-)instructions accurately and addressing hallucination issues.

\paragraph{Discussion on ``shortcut learning.''} 
We find interesting yet concerning prompt edits on the counterfactual task of base-8 addition. When induction initialization is used (\textit{i.e.}, PE2 is uninformed with the information of base-8 and must infer it from the examples), PE2 is able to devise its own heuristic that is partially correct (``... if both numbers are less than 50, add 2 to the sum. If either number is 50 or greater, add 22 to the sum''; see Table~\ref{table:case_study}).
This heuristic holds true for a subset of test cases like 75+7=104 and 5+6=13, but is ultimately not the intended solution.

On the positive side, it demonstrates PE2's ability to adapt in unseen scenarios and engage in sophisticated counterfactual reasoning.
However, it is concerning that models prompted with these self-induced shortcut solutions achieve a test accuracy of 37\% (average over 5 runs), which outperform models explicitly prompted to perform base-8 addition (test acc: 17\%/28\% before/after PE2 optimization). 
Shortcut learning \cite{geirhos2020shortcut} has been studied extensively for gradient-based optimization. Our experiments suggest that similar failure modes may be present in automatic prompt optimization. 

\input{tables/failure_edits}

\subsection{Additional Analysis}
Due to space limit, we summarize our other findings below and defer the details to Appendix \ref{app:additional_analysis}.
\paragraph{Effect of Initialization. (\S\ref{app:analysis_init})} (1) PE2 is able to recover from misleading or irrelevant prompt initializations, however the final prompt after optimization is worse than when using an instructive initialization.
(2) We experiment with induction initialization. In this case, PE2 is able to discover a high quality prompt \textit{from scratch} that matches with ``Let's think step by step'' on MultiArith.

\paragraph{Effect of Task Format/Difficulty. (\S\ref{app:analysis_format})} We experiment with using a generative format (\textit{i.e.}, generating the answer string) and a multi-choice format (\textit{i.e.}, selecting from given choices A/B/C/D) on the Date Understanding task in BIG-bench Hard. 
We observe that automatic prompt engineering methods has limited effect on the multi-choice format, but bring significant gains on the generative format.

\paragraph{Do optimized prompts generalize to other LLMs? (\S\ref{app:analysis_generalization})} 
We evaluate prompts optimized for \texttt{text-davinci-003} on other models such as \texttt{mpt-7b-instruct}, \texttt{yi-6b} and \texttt{mistral-7b-instruct}. 
We do not observe consistent cross-model generalization trends. 
This suggests that, although PE2 is a model-agnostic prompt optimization method (\textit{i.e.}, can be applied to various $\mathcal{M}_{task}$), the final optimized prompts are specific to the underlying task model.

\section{Related Work}

\paragraph{Automatic Prompt Engineering.} 
To alleviate the intensive trial-and-error efforts in manual prompt engineering, the research community has developed various strategies to automate this process with techniques such as incremental editing \citep{prasad-etal-2023-grips}, reinforcement learning \citep{deng-etal-2022-rlprompt, zhang2022tempera}, algorithmic search \citep{xu2022gps}, generating in-context demonstrations adaptively \citep{wan-etal-2023-better,wan-etal-2023-universal}, among others.
A line of work focus on meta-prompting LLMs themselves for automatic prompt engineering \citep{honovich-etal-2023-instruction, zhou2023large, pryzant-etal-2023-automatic}. 
In our work, we discuss potential limitations in these methods and subsequently introduce new meta-prompt components in PE2. 

\paragraph{Prompting LLMs for Complex Reasoning Tasks.} 
Recent research works suggest that LLMs can perform complex reasoning tasks, \textit{e.g.}, grade-school math problems \citep{cobbe2021training}. There are two major techniques to boost LLMs' performance on this: \textbf{(1) prompting methods} that guide the model to produce intermediate reasoning steps, either with few-shot demonstrations \citep{nye2021show, wei2022cot, yao2023react} or with zero-shot prompts \citep{kojima2022large}; \textbf{(2) self-reflection methods} that progressively guide the model to inspect its current output and refine it \citep{chen2023teaching, madaan2023self, paul2023refiner, kim2023language}.
At its core, prompt engineering is a language generation task requiring complex reasoning. Human prompt engineers usually examine the failure cases produced by the current prompt closely, make hypotheses, and compose a new prompt. In this work, we explore various prompting strategies when building an LLM-powered automatic prompt engineer.

\paragraph{Self-training and Self-improving for LLMs.}
Self-training refers to the technique of using a weak model to annotate input-label pairs and using them for further training \citep{Rosenberg2005SemiSupervisedSO}.
In the context of LLMs, STaR \citep{zelikman2022star} and Self-Improve \citep{huang2022large} show that employing LLMs to generate high-quality reasoning chains, followed by model fine-tuning on these chains, can significantly improve the model's reasoning capabilities.
In this work, we consider textual prompts as the ``parameters'' of LLMs, and we optimize these ``parameters'' with LLMs.
This may be categorized as a case of self-improving \citep{noah_goodman_blog}.
More discussion in Appendix~\ref{app:contemporary}.

\section{Conclusion}

In this paper, we introduced three meta-prompt components that lead to improved performance on automatic prompt engineering. The resulting method PE2 refines prompts written by human experts and surpasses established automatic prompt engineering baselines across various scenarios, notably on counterfactual tasks and a production application.
Through comprehensive analysis and case studies, we illustrate PE2's ability to make targeted prompt edits and generate high-quality prompts, and demonstrate its general applicability with different LLMs.

The challenge of prompt engineering a prompt engineer remains ongoing.
As highlighted in our case study, we believe improving the LLM's instruction following abilities and mitigating hallucination issues will be crucial for improving automatic prompt engineering. 
As the capabilities of LLM continue to evolve, their potential involvement in optimization or feedback loops necessitates a deeper empirical understanding of their failure modes \cite{pan2024feedback}, including shortcut learning discovered in this study. 
Looking ahead, we are also excited about applying PE2 to optimize its own meta-prompt in a self-referential way, in line with \citet{metz2020tasks, irie2022modern, fernando2023promptbreeder, zelikman2023self}.

\section*{Limitations}
Firstly, we opt for a relatively small prompt search budget ($T=3$, $m=4$, $n=4$; see \S\ref{ssec:budget}) due to cost considerations. 
In most of our experiments, the performance tends to plateau after $T=3$ optimization steps. However, it's important to consider that the use of the initialization prompt "let's think step by step" in many cases might introduce a potential confounding factor. This prompt could be already near-optimal, leading to fast convergence during prompt optimization.
Given the stochastic nature of natural language generation sampling and prompt optimization dynamics, it is possible that a larger prompt search budget or different experimental settings could yield new insights and conclusions.

Secondly, our study uses proprietary models such as \texttt{gpt-4} and \texttt{text-davinci-003}. (1) It raises reproducibility concerns as proprietary models may undergo upgrades or discontinuation over time. However, we believe the core concepts introduced in this paper is model-agnostic. This is supported by our experiments where we use two different sets of ($\mathcal{M}_{proposal}$, $\mathcal{M}_{task}$) (see \S\ref{ssec:budget}) and experiments using \texttt{Mistral-7B-Instruct-v0.2} model as the task model. 
 (2) It also raises concerns on data contamination, as the tasks and prompts included in our study may or may not have been part of the model's training data.

Lastly, apart from the three translation tasks in the Instruction Induction benchmark, our study predominantly focuses on tasks in English. We recognize the importance of inclusivity in language technology and acknowledge the need to extend our research to a multilingual setting in the future.

\subsubsection*{Acknowledgments}
We thank anonymous reviewers, members of USC NLP, and members of Microsoft Office of Applied Research for their valuable feedback. In particular, Qinyuan Ye would like to thank Mayee Chen, Zhuoran Lu, Onkar Kulkarni, Jakob Schoeffer, Connor Lawless and Marios Papachristou for the insightful conversations and for making her internship a truly memorable experience. Qinyuan Ye was supported by a USC Annenberg Fellowship.
\bibliography{reference, acl_anthology}
\newpage

\appendix
\newpage

\section{Additional Results and Analysis}\label{app:additional_analysis}

\subsection{Effect of Initialization} \label{app:analysis_init}
\begin{figure}[ht]
    \centering
    \includegraphics[width=0.48\textwidth, clip, trim=0 0 0 1.1em]{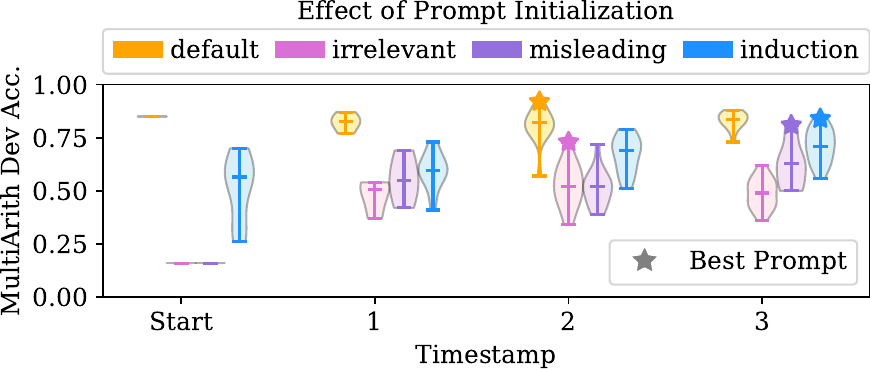}
    \caption{Prompt optimization dynamics on MultiArith when different prompt initializations are used.}
    \label{fig:initialization}
\end{figure}

Previously, we use ``Let's think step by step'' as the initialization for math reasoning tasks.
We further experiment with using a \textit{misleading} prompt, an \textit{irrelevant} prompt and \textit{induction} initialization (induction from a few examples). The results are presented in Table~\ref{tab:initialization} and the optimization dynamics are visualized in Fig.~\ref{fig:initialization}. 

\input{tables/initialization}

In general, performance drops when alternative initialization methods are used, which highlights the importance of high-quality initialization. Still, PE2 is able to override the irrelevant or misleading prompts and gradually improve the performance (Fig.~\ref{fig:initialization}).
Remarkably, PE2 is able to discover a high quality prompt\footnote{MultiArith prompt found by PE2 using induction initialization: ``Analyze the problem and perform the calculations. Consider addition, subtraction, division, multiplication and perform them in the order they appear. If required, round up results to the nearest whole number. Subtract done tasks from total when necessary.''} by itself using induction initialization (84\% on MultiArith-Dev) that almost matches with ``Let's think step by step'' (85\%) designed by highly-experienced human prompt engineers. This demonstrates the impressive prompt engineering capability of PE2 and suggests its potential for finding even better prompts when given additional computational resources.

\subsection{Effect of Task Format}\label{app:analysis_format}
For Date Understanding from BIG-bench Hard, we experiment with both a generative format (\textit{i.e}., generating the answer string; used in \citet{gao2023pal}) and a discriminative/multi-choice format (\textit{i.e.}, selecting from given choices A/B/C/D; used in \citet{suzgun-etal-2023-challenging}). For Movie Recommendation, we experiment with two different multi-choice formats.
See Table~\ref{tab:task_format} for the formats that we used.

\input{tables/task_format}
\begin{figure}[ht]
    \centering
    \includegraphics[width=0.48\textwidth]{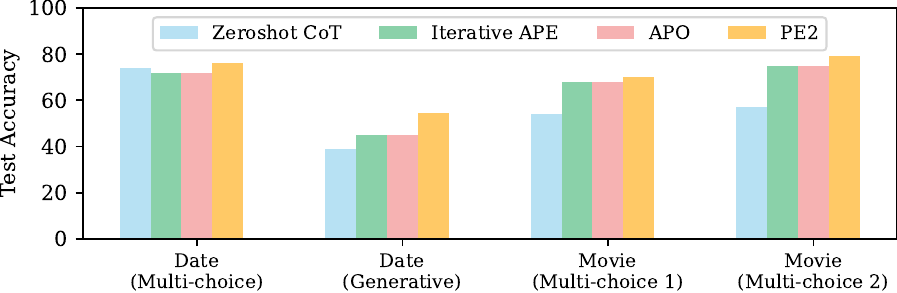}
    \caption{Effect of Task Format. See Table~\ref{tab:task_format} for the formats used.}
    \label{fig:task_format}
\end{figure}

We report the results in Fig.~\ref{fig:task_format}. For Date Understanding, the multi-choice format narrows the output space and thus lower the difficulty of the task. We hypothesize that in combination with Zero-shot CoT, the task performance is close to saturation and automatic prompt engineering method does not provide extra benefit. The task is more challenging in the generative format and the detailed instructions in the optimized prompt bring significant performance gains. 

For Movie Recommendation, we found that prompt optimization methods bring significant performance gains in both cases. The optimized prompts contain multi-step plans (\textit{e.g.}, consider genre, director, ...), which boost the task performance. Minor formatting decisions such as outputting the option letter or the option string can still mildly affect the final accuracy.

Overall, the question of ``when is automatic prompt optimization most effective'' is dependent on many factors, including task format, task difficulty, output space size, task model's instruction following abilities, etc.

\subsection{Do optimized prompts generalize to other LLMs?} \label{app:analysis_generalization}

\begin{figure}[ht]
    \centering
    \includegraphics[width=0.48\textwidth]{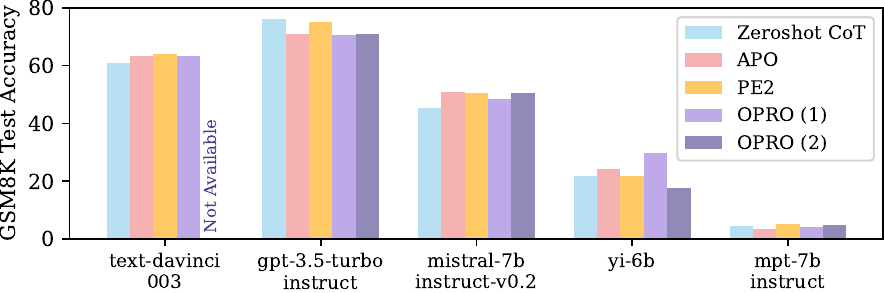}
    \caption{Analysis on generalizability of prompts across models. See Table~\ref{tab:gsm8k_prompts} for the prompts used in this study.}
    \label{fig:transferability}
\end{figure}

We evaluate 5 GSM8K prompts for our prompt generalization study (see Table~\ref{tab:gsm8k_prompts}). Note that the APO and PE2 prompts are optimized for \texttt{text-davinci-003}. The two prompts reported in OPRO \cite{yang2023large} are optimized for \texttt{PaLM 2-L}.
We investigate the generalization of optimized prompts by evaluating them on four models: \texttt{gpt-3.5-turbo-instruct}, \texttt{mistral-7b-instruct-v0.2}, \texttt{yi-6b}, and \texttt{mpt-7b-instruct}. We report the results in Fig.~\ref{fig:transferability}.

\input{tables/gsm8k_prompts}

Our results do not exhibit consistent generalization trends. Optimized prompts generally outperforms the original zero-shot CoT prompt on \texttt{text-davinci-003}, \texttt{mistral-7b-instruct-v0.2}, \texttt{yi-6b}. 
However, with \texttt{gpt-3.5-turbo-instruct}, the original CoT prompt outperforms all optimized prompts. 
Our hypothesis is that ``Let's think step by step'' is included in public instruction tuning collections \cite{kim-etal-2023-cot} and thus models trained on these collections may perform better with this special prompt.
However the instruction tuning mixture used for training \texttt{gpt-3.5-turbo-instruct} are not disclosed, and thus we cannot further investigate this.

Overall, our results suggest that current automatic prompt optimization methods tend to find model-specific prompts that \textit{do not} reliably generalize to alternative models. 
This conclusion contrasts with the findings in PromptAgent \cite{wang2023promptagent}, which we attribute to discrepancies in experimental setup. To maintain consistency with prior research \cite{zhou2023large}, we have limited the prompt length to be 50 or 200 tokens. The conclusion may differ when this constraint is removed.

Future work may develop robust prompt optimization methods that operate across multiple task models, in a way similar to \citet{li-etal-2023-robust} which operates across domains. This may help identify high-quality prompts invariant to the underlying task model, so that when new and more powerful models (\textit{e.g.}, GPT-5) are released, the optimized prompt may be used directly.

\vspace{1cm}
\begin{flushright}
\textbf{(Continued on next page)}
\end{flushright}

\onecolumn
\section{Other Meta-Prompt Components We Tried}\label{app:meta-prompt-optim}
In addition to the meta-prompt components studies in the main paper, we also tried other components in the early stage of PE2's development. As the results are mixed and inconclusive on these components, we report them here in the appendix. We illustrate these components in Fig.~\ref{fig:intro-old}.

\begin{figure*}[ht]
    \centering
    \includegraphics[width=\textwidth]{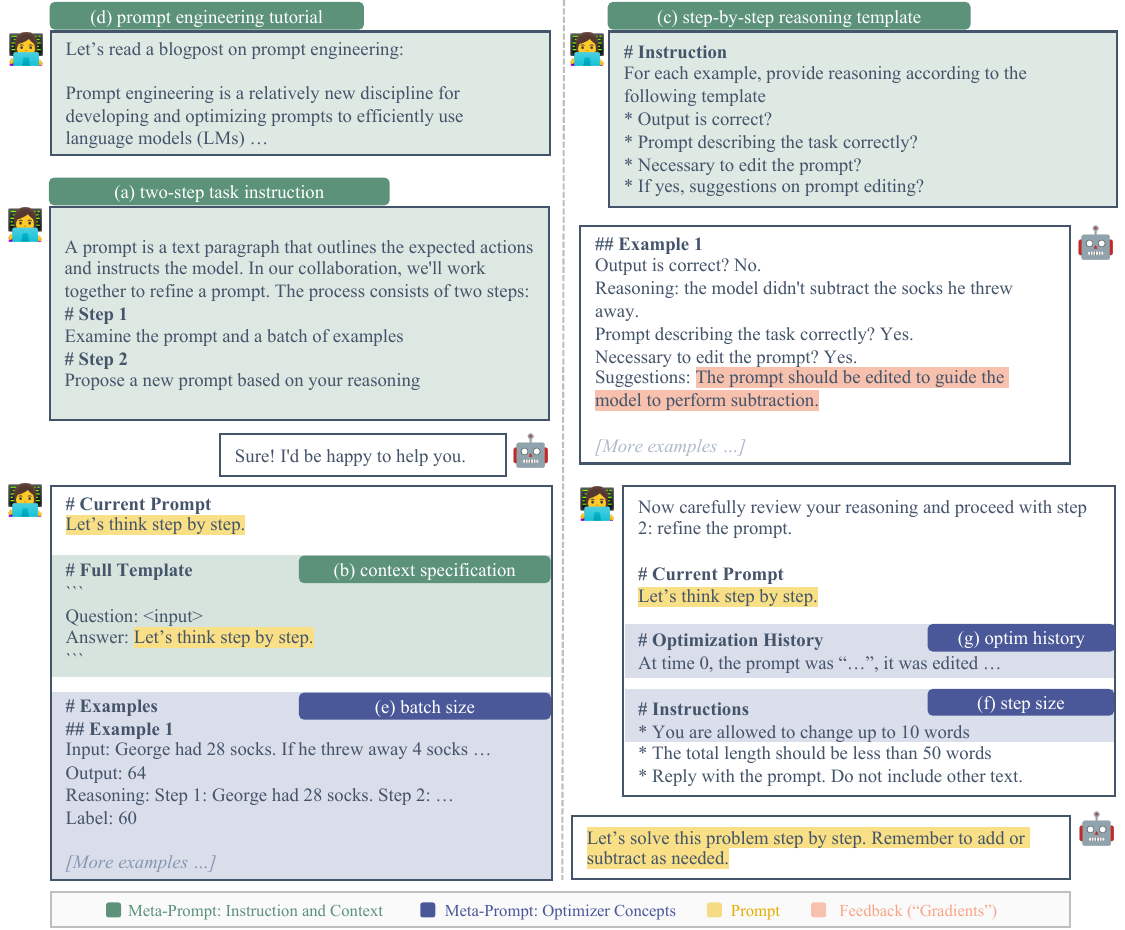}
    \caption{Illustration of meta-prompt components discussed in Appendix~\ref{app:meta-prompt-optim}.}
    \label{fig:intro-old}
\end{figure*}

\paragraph{Providing Detailed Instructions and Context. } 
\begin{enumerate}[label=(\alph*), left=0pt, start=4]

    \item \textbf{Prompt Engineering Tutorial.} To help the LLM better understand the task of prompt engineering, we provide an online tutorial of prompt engineering in the meta-prompt.\footnote{\url{https://www.promptingguide.ai/introduction}. Published under MIT license.}
    \end{enumerate}

\paragraph{Incorporating Common Optimizer Concepts.} 
The prompt engineering problem described in Eq.~\ref{eq:pe} is essentially an optimization problem, 
and the prompt proposal in Eq.~\ref{eq:optim} can be considered as doing one optimization step. 
Thus, we consider the following concepts commonly used in gradient-based optimization and develop their \textit{verbalized counterparts} to be used in our meta-prompt.
\begin{enumerate}[label=(\alph*), left=0pt, start=5]
    \item \textbf{Batch Size.} Batch size is the number of (failure) examples that is used in each prompt proposal step (Eq.~\ref{eq:optim}). 
    By default PE2 uses a batch size of 2. 
    We experiment with batch sizes of $\{1,4,8\}$ additionally in this section. 
    \item \textbf{Step Size.} In gradient-based optimization, the step size determines the extent to which the model's weights are updated. In prompt engineering, the counterpart would be the number of words (tokens) that can be modified. We directly specify that ``You are allowed to change up to $s$ words in the original prompt'' in the meta-prompt, where $s\in\{5, 10, 15, \text{None}\}$.\footnote{\citet{chen2022controllable} and \citet{pmlr-v202-zhou23g} showed that LLMs could follow text generation constraints specified in natural language.} 
    \item \textbf{Optimization History and Momentum.} 
    Momentum \citep{qian1999momentum} is a technique to accelerate optimization and avoid oscillations by maintaining the moving average of past gradients. To develop the verbalized counterpart of momentum, we include all past prompts (at timestamp $0, 1, ..., t-1$), their performance on the dev set, and a summary of prompt edits.
\end{enumerate}

\input{tables/ablation2}

\paragraph{Results and discussion.} 
We report the results when these components are used in Table~\ref{table:ablation2}.
We do not observe significant improvement by incorporating prompt engineering tutorial. As the tutorial is excessively long (2500+ tokens) and slows down the runtime, we do not include it in the final version of PE2. 
The optimizer-inspired concepts can improve the performance occasionally, but the current experiments do not give a consistent conclusion regarding their utilities.

Similar to the challenges encountered in gradient-based optimization, the process of hyperparameter selection is inherently noisy and often varies depending on the task at hand.
For discrete prompt optimization, this complexity is further compounded by factors such as the task model's sensitivity to prompts and the proposal model's capability to follow instructions in the meta-prompt. 
For example, \citet{sun-etal-2023-evaluating} point out that LLMs struggled at meeting fine-grained constraints such as ``generate exactly 5 words,'' which could potentially diminish the effectiveness of the (f) step size component.
Additionally, (g) momentum requires multiple optimization steps to accumulate, yet our experiments are restricted to three steps due to cost constraints.

Although these meta-prompt components do not currently take effect consistently with existing LLMs and experimental settings, their potential justifies re-examination in the future, particularly as models become more capable, and the efficiency and scalability of automatic prompt engineering methods improve.

\section{Additional Discussion}
\subsection{Recent Works}\label{app:contemporary}

A recent work \citep{yang2023large} introduced the concept of large language models as optimizers and proposed optimization by prompting (OPRO).
In the following, we discuss the differences and connections between OPRO and our work.

\noindent\textbf{(1) Focus of the work.} Both OPRO and our work conduct experiments on prompt optimization; the focus of the two works differ.
OPRO can be applied to general optimization problems, including linear regression and traveling salesman problem. In our work we limit the scope to prompt optimization, with a specific focus on proposing and investigating different components in the meta-prompt.

\noindent\textbf{(2) Optimization strategy.} The optimization strategies of the two works are different. PE2 is largely inspired by the concepts in APO \citep{pryzant-etal-2023-automatic}, instructing the model to produce textual feedback (``gradient'') \textit{explicitly}. It is more analogous to gradient descent. 
OPRO uses the execution accuracy as rewards to guide the optimization \textit{indirectly}, which, in our understanding, is more analogous to in-context RL methods \citep{shinn2023reflexion}. For future work, it would be interesting to compare the effectiveness and efficiency of both methods in a controlled setup.

\noindent\textbf{(3) Challenges in making direct comparison.}
\citet{yang2023large} mainly uses \texttt{PaLM 2-L} model and \texttt{text-bison} model as the task model (scorer), and optimizes the prompt for up to 200 steps.
In our work, we mainly use \texttt{text-davinci-003} and \texttt{GPT-4}, and optimize the prompt for 3 steps by default. Due to access and budget constraints, we are unable to make direct comparison with OPRO.

In addition to OPRO,  several recent works have explored automatic prompt engineering using diverse strategies. PromptBreeder \cite{fernando2023promptbreeder} adopts a self-referential prompt evolution framework, employing mutation prompts (similar to the concept of ``meta-prompts'' discussed in this paper) to edit task prompts, and hyper-mutation prompts to edit mutation prompts. PromptAgent \cite{wang2023promptagent} adopts the Monte Carlo Tree Search algorithm that iteratively performs selection, expansion, simulation and back-propagation for strategic prompt editing. Evoke \cite{hu2024evoke} introduces a collaborative approach where an LLM-reviewer and an LLM-author work together to refine the prompt using critical thinking. In parallel to these works, we focus on the design and evaluation of the \textit{meta-prompt} in LLM-powered automatic prompt engineering in this paper.

Our work is also related to Self-Discover \cite{zhou2024self}, a framework for LLMs to compose reasoning structures, such as ``break down into sub-tasks'' for complex tasks. PE2 demonstrates task decomposition behaviors as discussed in \S\ref{ssec:pe2_behavior} and Table~\ref{table:case_study}, which can be seen as presence of rudimentary meta-reasoning capabilities in LLMs.

\subsection{Discussion on using PE2 to optimize its own meta-prompt}
Conceptually, PE2 may be applied to not only optimize prompts, but also meta-prompts. We can replace $p^{(t)}$ and $p^{(t+1)}$ in Eq.~\ref{eq:optim} with the meta-prompt $p^{meta}$ directly to enable PE2 to optimize the meta-prompt. We believe this is an exciting direction to pursue. However, three challenges (and broader questions) arise if we pursue this direction, and we look forward to addressing these challenges in the future:
\begin{enumerate}[noitemsep,topsep=0pt,leftmargin=14pt]
    \item How to collect data for such a study? To ensure this meta-prompt is general we may need a large collection of tasks along with prompt optimization history associated with them. Creating a resource like this will be a large effort.
    \item How to automatically optimize the meta-prompt when there are no ground truth labels for prompt engineering? Math problems have ground-truth answers so that PE2 can inspect them and provide feedback for prompt refinement. The task of prompt engineering does not have ground truth labels, and this potentially makes the meta-prompt optimization process more noisy.
    \item It would be very costly to run and even evaluate a system like this. To \textit{evaluate} one meta-prompt candidate and show it outperforms other meta-prompt candidates, we will need to use it for prompt optimization on various tasks. We would expect the \textit{optimization} process of the meta-prompt to be a magnitude more costly.
\end{enumerate}

\section{Additional Experiment Details}\label{app:exp_details}
\subsection{Prompt Search Algorithm}
See Algorithm~\ref{algo:search}.

{\begin{figure*}[h]
\centering
\begin{minipage}{.9\linewidth}
\begin{algorithm}[H]
\small
\caption{Search Procedure}\label{algo:search}
\begin{algorithmic}[1]  
\State $P^{(0)}=P_{init}$ or $P^{(0)}=\mathcal{M}_{init}(x_1, y_1, ..., x_n, y_n;p^{init})$ \hfill\Comment{Manual init. or induction init.}
\For{$t=0, ..., T-1$}
    \State $P^{(t+1)} = \emptyset$
    \For{$p^{(t)}\in \text{Select-Best}(\cup_{i=0}^t P^{(i)}, n)$}\hfill\Comment{Select best $n$ prompts based on $D_{dev}$}
        \For{$j=1...m$}
            \State $B = \text{Sample}(D_{train})$\hfill\Comment{Sample a batch (random or failure examples)}
            \State $p^{(t+1)} = \mathcal{M}_{optim}(p^{(t)}, B; p^{meta})$\hfill\Comment{New prompt proposal}
            \State $P^{(t+1)} = P^{(t+1)} \cup \{p^{(t+1)}\}$
        \EndFor
    \EndFor    
\EndFor 
\State \Return Select-Best$(\cup_{i=0}^T P^{(i)}, 1)$\hfill\Comment{Return the final best prompt based on $D_{dev}$}
\end{algorithmic}
\end{algorithm}
\end{minipage}
\end{figure*}
}

\subsection{Controlling Prompt Length}
By default the max length of prompts is set to be 50 tokens, following \citet{zhou2023large}. For counterfactual tasks, to allow more space to explain the counterfactual situations, the max length is set to be 200 tokens.

\subsection{Infrastructure and Runtime}
\paragraph{Infrastructure.} We use OpenAI API\footnote{\url{https://openai.com/blog/openai-api}} to access \texttt{text-davinici-003}, \texttt{gpt-3.5-turbo-instruct}, \texttt{gpt-4}, \texttt{gpt-4-turbo}. For prompt generalization experiments using \texttt{mistral-7b-instruct}, \texttt{mpt-7b-instruct} and \texttt{yi-6b}, we run experiments locally using one Nvidia RTX A6000 GPU and the vLLM toolkit \cite{kwon2023efficient}.

\paragraph{Runtime.} One prompt optimization experiment using  \texttt{gpt-4}/\texttt{text-davinici-003} as task/proposal model takes about 90 minutes. This is also subject to API rate limits.

\paragraph{Costs.} When using \texttt{gpt-4}/\texttt{text-davinci-003} it costs about \$25 USD for one prompt optimization experiment. In the later stage of this project, we use \texttt{gpt-4-turbo}/\texttt{gpt-3.5-turbo-instruct} which are newer and cheaper, and the cost is reduced to about \$3 USD per experiment. 

\subsection{Tasks and Data}\label{app:data}
We summarize the dataset size and data split information in Table~\ref{tab:dataset}. We summarize the source and license information of the datasets in Table~\ref{tab:license}. To the best of our knowledge, our usage of these datasets are consistent with their intended use; the data we use do not contain personal or sensitive information. Most of the datasets are in English and not domain-specific.
\begin{table*}[ht]
\centering
\scalebox{0.75}{
\begin{tabular}{llllll}
\toprule
Dataset                     & Subtasks & $|T_{train}|$ & $|T_{dev}|$ & $|T_{test}|$ & \# Random Samples \\\midrule
MultiArith \citep{roy-roth-2015-solving} & - & 100 & 100 & 400 & 1 \\
GSM8K \citep{cobbe2021training} & - & 100 & 100 & 1319 & 1  \\
Instruction Induction \citep{honovich-etal-2023-instruction}        & 14 Subtasks & 100 & 20 & 100 & 5   \\
Counterfactual Eval \citep{wu2023reasoning} & 12 Subtasks & 100 & 20 & 100 & 5 \\
BIG-Bench Hard (BBH format used in \citet{suzgun-etal-2023-challenging}) & 27 Subtasks & 100 & 100 & 50 & 1 \\
BIG-Bench Hard (Alternative format; see \S\ref{app:analysis_format}) & 2 Subtasks & 100 & 100 & 500 & 1 \\\bottomrule 
\end{tabular}
}
\caption{Dataset sizes and data splits.}\label{tab:dataset}
\end{table*}
\begin{table*}[ht]
\centerline{
\scalebox{0.75}{
\begin{tabular}{lll}
\toprule
Dataset                     & License & Source \\\midrule
MultiArith \citep{roy-roth-2015-solving} & Unknown & \url{https://github.com/wangxr14/Algebraic-Word-Problem-Solver/} \\
GSM8K \citep{cobbe2021training} & MIT & \url{https://github.com/openai/grade-school-math}  \\
Instruction Induction \citep{honovich-etal-2023-instruction}        & Apache-2.0 & \url{https://github.com/orhonovich/instruction-induction}   \\
Counterfactual Eval \citep{wu2023reasoning} & Unknown & \url{https://github.com/ZhaofengWu/counterfactual-evaluation} \\
BIG-bench Hard \citep{suzgun-etal-2023-challenging} & Apache-2.0 & \url{https://github.com/google/BIG-bench} (original) \\
& & \url{https://github.com/suzgunmirac/BIG-Bench-Hard} (reformatted) \\
\bottomrule 
\end{tabular}
}
}
\caption{License and Source of the datasets used in this study.}\label{tab:license}
\end{table*}

\paragraph{(1) Mathematical Reasoning.} 
The MultiArith dataset \citep{roy-roth-2015-solving} contains 600 examples. As our prompt optimization method requires a training set, we randomly split into 100/100/400 for train/dev/test. This creates a slight discrepancy when comparing the results with past reported results. We ensure our reproduction is fair across different methods by using this fixed split.
The GSM8K dataset \citep{cobbe2021training} has a provided test split (1319 examples). We randomly selected 200 examples for the original train split, and use 100 as $D_{train}$ and 100 as $D_{dev}$.

\paragraph{(2) Instruction Induction.} We closely follow the settings in \citet{zhou2023large}. For each subtask, we randomly sample 5 different $D_{train}/D_{dev}/D_{test}$ of size 100/20/100. We list the sub-tasks in Instruction Induction benchmark in Table~\ref{tab:ii}. We removed 8 tasks (active to passive, diff, first word letter, letters list, num to verbal, singular to plural, sum), because our baseline method APE \cite{zhou2023large} already achieves near perfect accuracies (95\%+) on these tasks. We also removed 2 tasks (cause and effect, common concept) because they have less than 50 examples in total, and it is challenging to create train/dev/test split from these examples.

\input{tables/ii}

\paragraph{(3) BIG-bench Hard Tasks.} 
We mainly experiment with the BBH task format used in \citet{suzgun-etal-2023-challenging}. As the public BBH repository have 250 examples per task, we randomly split them into 100/100/50 for $D_{train}/D_{dev}/D_{test}$.
For Date Understanding and Movie Recommendation, we consider using alternative tasks formats to study the their effect (see \S\ref{app:analysis_format}). We obtain the data from the original BIG-bench repository which contains more examples per task. 
Hence we randomly sample 100/100/500 examples for $D_{train}/D_{dev}/D_{test}$ in these two experiments. 

\paragraph{(4) Counterfactual Evaluation.} We use three subtasks in this evaluation suite: arithmetic, chess and syntax. For each subtask, we randomly sample 5 different $D_{train}/D_{dev}/D_{test}$ of size 100/20/100. We list the sub-tasks in Table~\ref{tab:cf}.

\input{tables/cf}

\paragraph{(5) Production Prompt.} We use a randomly sampled subset of human annotated queries and labels ($>150$), which are derived from user reported errors. The data is divided between training ($50\%$), validation ($25\%$) and testing ($25\%$).  We use the F1-score for evaluating model outputs and report the absolute change in score with the initialization prompt.

\newpage
\onecolumn
\section{Additional Result Figures and Tables}\label{app:additional_results}

\paragraph{Notable Prompt Edits.} Additional examples on notable prompt edits made by PE2 are in Table~\ref{table:case_study2}.
\input{tables/successful_edits2}

\begin{figure*}[ht]
    \centering
    \includegraphics[width=\textwidth]{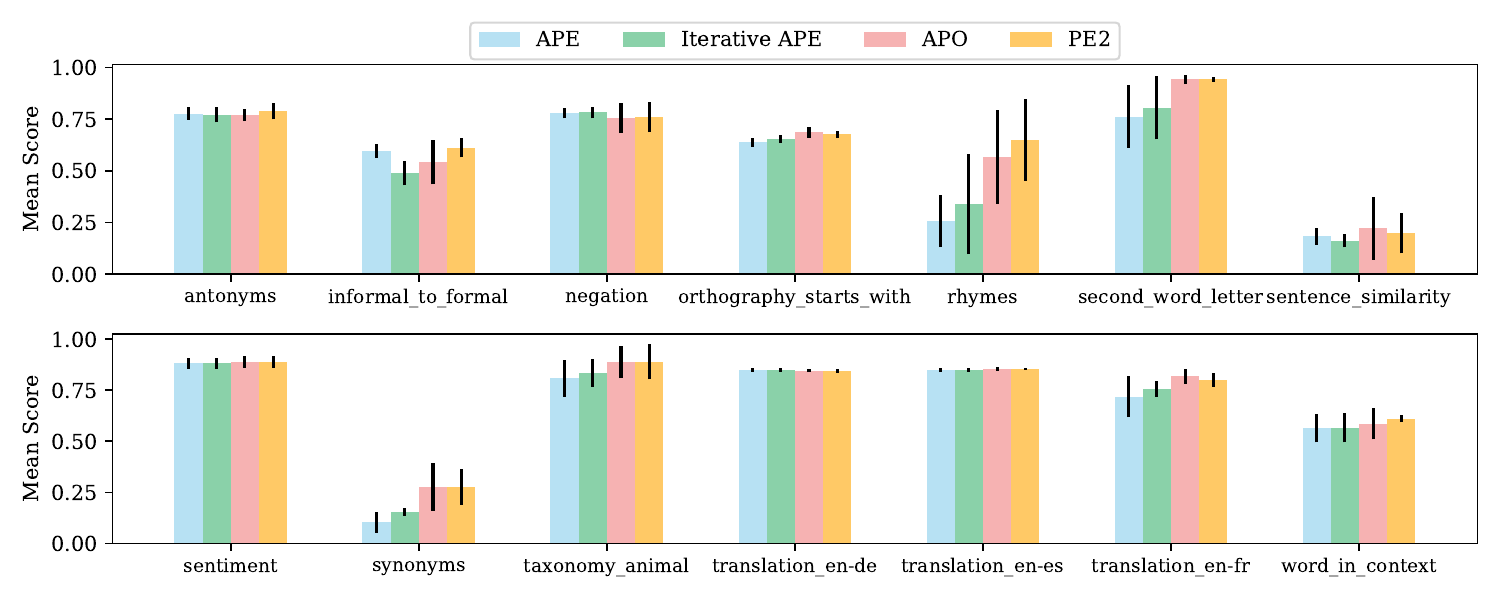}
    \caption{Results on the Instruction Induction Benchmark. The performance of APO and PE2 are close to each other on most tasks. Our hypothesis is that tasks in Instruction Induction Benchmark are relatively easier compared to the other benchmarks, leading to performance saturation. Raw results in Table~\ref{table:ii_raw}.}
    \label{fig:ii} 
\end{figure*}

\begin{figure*}[ht]
    \centering
    \includegraphics[width=\textwidth]{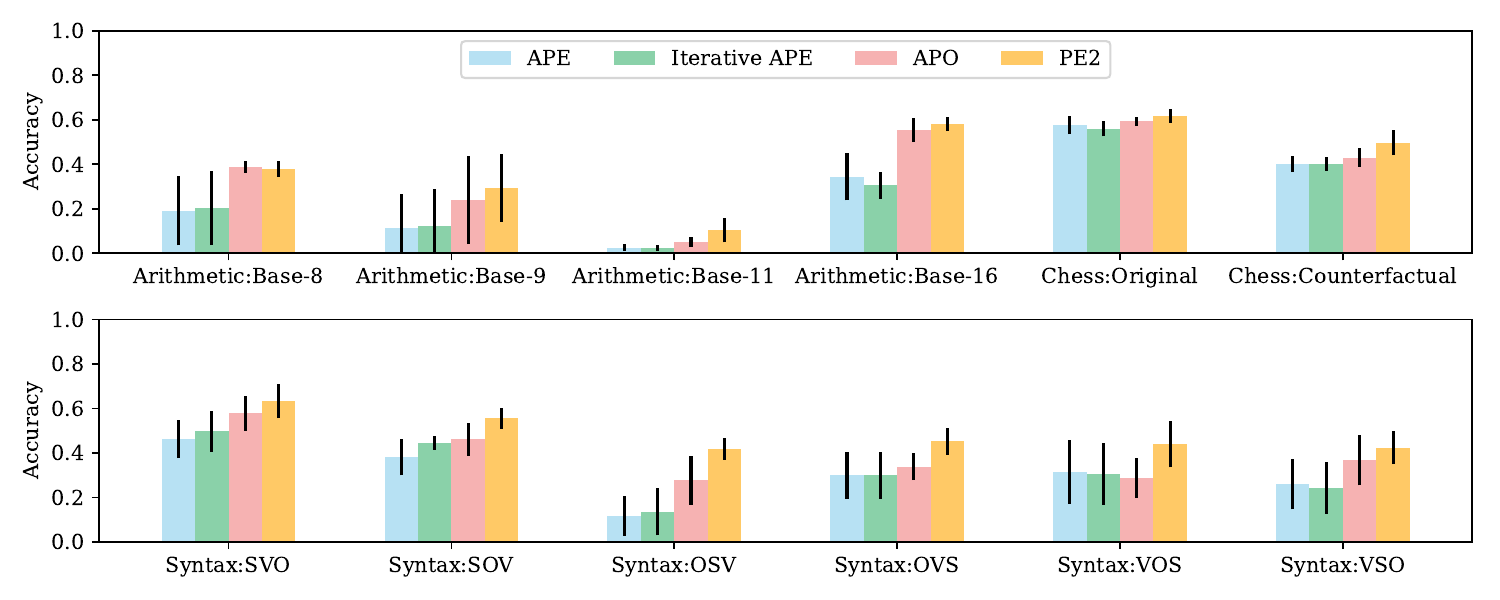}
    \caption{Results on Counterfactual Eval (Induction Initialization). Raw results in Table~\ref{table:cf_raw1}.
    }
    \label{fig:results-cf}
\end{figure*}

\begin{figure*}[ht]
    \centering
    \includegraphics[width=\textwidth]{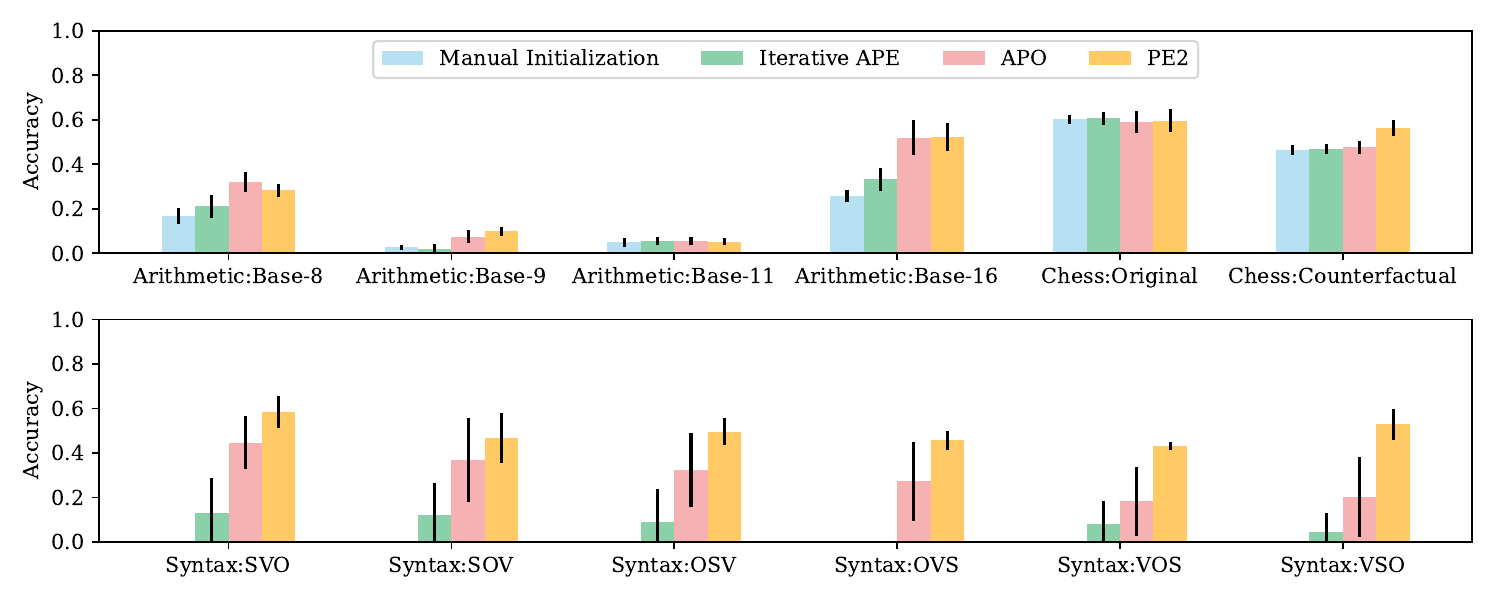}
    \caption{Results on Counterfactual Eval (Manual Initialization). Raw Results in Table~\ref{table:cf_raw2}.}
    \label{fig:cf_init}
\end{figure*}

\begin{figure*}[ht]
    \centering
    \includegraphics[width=\textwidth]{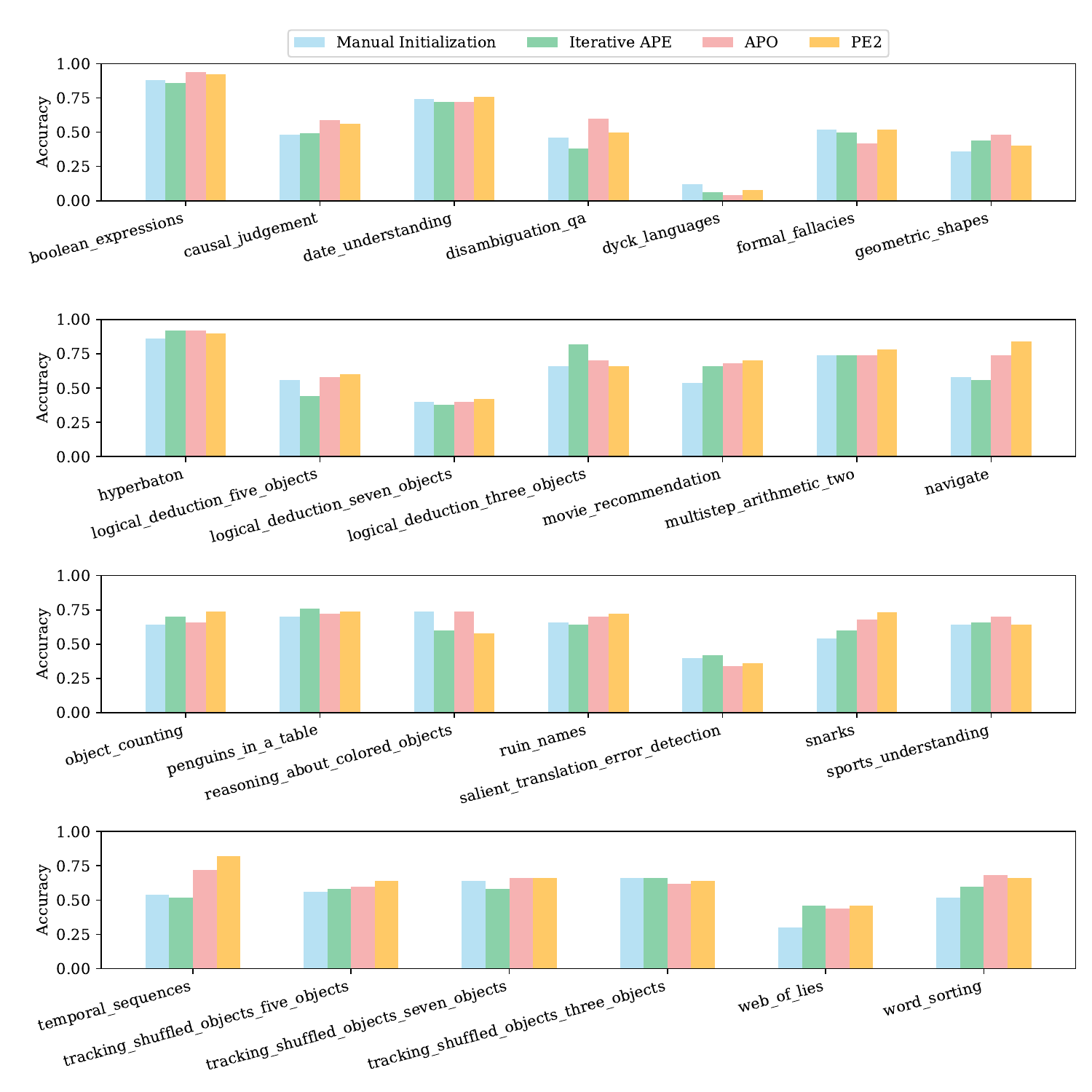}
    \caption{Results on the BIG-bench Hard \cite{suzgun-etal-2023-challenging}. Raw Results in Table~\ref{table:raw_bbh}.}
    \label{fig:bbh}
\end{figure*}

\paragraph{Results Breakdown.} 
We report the results on each subtask in Instruction Induction in Fig.~\ref{fig:ii} and Table~\ref{table:ii_raw}.
For counterfactual tasks, results using induction initialization are in Fig.~\ref{fig:results-cf} and Table~\ref{table:cf_raw1}; results using manual initialization are in Fig.~\ref{fig:cf_init} and Table~\ref{table:cf_raw2}. 
For BIG-bench Hard tasks, we report the results in Fig.~\ref{fig:bbh} and Table~\ref{table:raw_bbh}.
We report the results on Date Understanding and Movie Recommendation when alternative task formats are used in Table~\ref{table:bbh}.


\clearpage
\input{tables/raw_results_ii}
\input{tables/raw_results_counterfactual}
\input{tables/raw_results_bbh}

\newpage
\newpage
\clearpage

\section{Meta-prompts}
We implement the meta-prompts using the guidance toolkit\footnote{\url{https://github.com/guidance-ai/guidance}}, which enables multi-round conversations and supports basic handlebars-style syntax to control the workflow.
\label{app:meta-prompt}
\subsection{Initialization Prompt $p^{init}$}
\label{app:meta-prompt-init}
The initialization prompt is originally from APE \citep{zhou2023large}. In this paper, it is shared by all methods (Iterative APE, APO and PE2).

\begin{lstlisting}[language=json]
{{#system~}}
You are a helpful assistant.
{{~/system}}
                                           
{{#user~}}
I gave a friend an instruction and {{n_demo}} inputs. The friend read the instruction and wrote an output for every one of the inputs.
Here are the input-output pairs:

{{demos}}

What was the instruction? It has to be less than {{max_tokens}} tokens.
{{~/user}}

{{#assistant~}}
The instruction was {{gen 'instruction' [[GENERATION_CONFIG]]}}
{{~/assistant}}
\end{lstlisting}

\subsection{APE}

\begin{lstlisting}[language=json]
{{#system~}}
You are a helpful assistant.
{{~/system}}
                                           
{{#user~}}
Generate a variation of the following instruction while keeping the semantic meaning.

{{prompt}}

The new instruction has to be less than {{max_tokens}} words.
Reply with the new instruction. Do not include other text.
{{~/user}}

{{#assistant~}}
{{gen 'new_prompt' [[GENERATION_CONFIG]]}}
{{~/assistant}}
\end{lstlisting}

\subsection{APO}

Part 1 - Generating ``gradients''

\begin{lstlisting}[language=json]
{{#system~}}
You are a helpful assistant.
{{/system~}}

{{#user~}}
I'm trying to write a zero-shot classifier prompt.

My current prompt is:
"{{prompt}}"

But this prompt gets the following examples wrong:
{{failure_string}}

Give {{n_reasons}} reasons why the prompt could have gotten these examples wrong. Do not include other text.
{{/user~}}

{{#assistant~}}
{{gen 'gradients' temperature=0.0}}
{{/assistant~}}
\end{lstlisting}

Part 2 - Refining the prompt

\begin{lstlisting}[language=json]
{{#system~}}
You are a helpful assistant.
{{/system~}}

{{#user~}}
I'm trying to write a zero-shot classifier.

My current prompt is:
"{{prompt}}"

But it gets the following examples wrong:
{{failure_string}}

Based on these examples the problem with this prompt is that:
{{gradient}}

Based on the above information, I wrote an improved prompt. The total length of the prompt should be less than {{max_tokens}} words.
{{/user~}}

{{#assistant~}}
The improved prompt is {{gen 'new_prompt' temperature=0.0}}
{{/assistant~}}
\end{lstlisting}

\subsection{PE2}
\label{app:pe2_meta_prompt}

\begin{lstlisting}[language=json]
{{#system~}}
You are a helpful assistant.
{{~/system}}

{{#if instruction}}
{{#user~}}
Let's read a blogpost on prompt engineering:
{{instruction}}
{{~/user}}
{{/if}}

{{#user~}}
A prompt is a text paragraph that outlines the expected actions and instructs the model to generate a specific output. This prompt is concatenated with the input text, and the model then creates the required output.

In our collaboration, we'll work together to refine a prompt. The process consists of two main steps:

## Step 1
I will provide you with the current prompt, how the prompt is concatenated with the input text (i.e., "full template"), along with {{batch_size}} example(s) that are associated with this prompt. Each examples contains the input, the reasoning process generated by the model when the prompt is attached, the final answer produced by the model, and the ground-truth label to the input. Your task is to analyze the examples, determining whether the existing prompt is decsribing the task reflected by these examples precisely, and suggest changes to the prompt.

## Step 2
Next, you will carefully review your reasoning in step 1, integrate the insights to craft a new, optimized prompt. Optionally, the history of refinements made to this prompt from past sessions will be included. Some extra instructions (e.g., the number of words you can edit) will be provided too.
{{~/user}}
                    
{{#assistant~}}
Sure, I'd be happy to help you with this prompt engineering problem. 
Please provide me with the prompt engineering history, the current prompt, and the examples you have.
{{~/assistant}}

{{#user~}}
## Prompt
{{prompt}}

## Full Template
This describes how the prompt of interested is concatenated with the input text. 
The prompt may appear before the input text, or after the input text.
Optionally the full template may contain other template information.
```
{{full_prompt}}
```

## Examples
{{examples}}

## Instructions
For some of these examples, the output does not match with the label. This may be due to the prompt being misleading or not describing the task precisely.

Please examine the example(s) carefully. Note that the ground-truth labels are __absolutely correct__, but the prompts (task descriptions) may be incorrect and need modification. For each example, provide reasoning according to the following template:

### Example <id>
Input: <input>
Output: <output>
Label: <label>
Is the output correct compared to the label: <yes or no, and your reasoning>
Is the output correctly following the given prompt: <yes or no, and your reasoning>
Is the prompt correctly describing the task shown by the input-label pair: <yes or no, and your reasoning>
To output the correct label, is it necessary to edit the prompt: <yes or no, and your reasoning>
If yes, provide detailed analysis and actionable suggestions to edit the prompt: <analysis and suggestions>
{{~/user}}

{{#assistant~}}
{{gen 'reasoning' temperature=0}}
{{~/assistant}}

{{#user~}}
Now please carefully review your reasoning in Step 1 and help with Step 2: refining the prompt. 

{{#if history}}
## Prompt Refinement History from the Past
Note that higher accuracy means better. If some edits are useful in the past, it may be a good idea to make edits along the same direction.
{{history}}
{{/if}}

## Current Prompt
{{prompt}}

## Instructions
{{#if step_size}}
* You are allowed to change up to {{step_size}} words in the original prompt.
{{/if}}
{{#if max_tokens}}
* The total length of the prompt should be less than {{max_tokens}} words.
{{/if}}
* Please help edit the prompt so that the updated prompt will not fail on these examples anymore.
* Reply with the prompt. Do not include other text.
{{~/user}}

{{#assistant~}}
{{gen 'new_prompt' temperature=0.7 max_tokens=300}}
{{~/assistant}}

{{#if history}}
{{#user~}}
Now please summarize what changes you've made to the prompt, in the following format. Make sure the summariy is concise and contains no more than 200 words.

" * At step {{timestamp}}, the prompt has limitations such as <summary of limitations>. Changes to the prompt include <summary of changes>."

Reply with the summarization. Do not include other text.
{{~/user}}

{{#assistant~}}
{{gen 'new_history' temperature=0.7 max_tokens=200}}
{{~/assistant}}
{{/if}}
\end{lstlisting}

\section{Prompt Optimization Results}
See Table~\ref{tab:all_prompts_start}-\ref{table:mistral-prompts}.
{
\scriptsize
\clearpage
\input{tables/prompts1}
\input{tables/prompts2}

\input{tables/prompts3}
\input{tables/mistral_prompts}
}
\end{document}

%% file: math_commands.tex
\usepackage{amsmath,amsfonts,bm}

\def\eqref#1{equation~\ref{#1}}

\def\1{\bm{1}}

\DeclareMathAlphabet{\mathsfit}{\encodingdefault}{\sfdefault}{m}{sl}
\SetMathAlphabet{\mathsfit}{bold}{\encodingdefault}{\sfdefault}{bx}{n}

\DeclareMathOperator*{\argmax}{arg\,max}

%% file: tables/math_results.tex
\begin{table}[t]
\scalebox{0.72}{
\begin{tabular}{l|cc|cc}
\toprule
\textbf{Method}        & \multicolumn{1}{c}{\begin{tabular}[c]{@{}c@{}}\textbf{Task}\\ \textbf{Model}\end{tabular}} & \multicolumn{1}{c}{\begin{tabular}[c]{@{}c@{}}\textbf{Proposal}\\ \textbf{Model}\end{tabular}} & \multicolumn{1}{|c}{\begin{tabular}[c]{@{}c@{}}\textbf{MultiArith}\\ \textbf{Test}\end{tabular}} & \multicolumn{1}{c}{\begin{tabular}[c]{@{}c@{}}\textbf{GSM8K}\\ \textbf{Test}\end{tabular}}\\\midrule
\rowcolor{gray!20}\multicolumn{5}{l}{Fixed Prompt, Reported by \citet{zhou2023large}}                                  \\\midrule
Zero-shot CoT & TD002 & - & 78.7 & 40.7 \\
APE & TD002 & TD002 & 82.0 & 43.0 \\\midrule
\rowcolor{gray!20}\multicolumn{5}{l}{Fixed Prompt, Reproduced}                                \\\midrule
Zero-shot CoT & TD003 & - & 86.0 & 60.9 \\
APE & TD003 & - & 87.0 & 61.5 \\\midrule
\rowcolor{gray!20}\multicolumn{5}{l}{Prompt Optimization}               \\\midrule
Iterative APE& TD003 & GPT-4 & 88.5 & 62.7 \\
APO & TD003 & GPT-4 & 88.5 & 63.1 \\
PE2 (this work) & TD003 & GPT-4 & \textbf{92.3} & \textbf{64.0} \\\bottomrule
\end{tabular}
}
\centering
\caption{Performance Comparison on Math Reasoning Tasks. TD002/003 stand for \texttt{text-davinci-002/003.} See Table~\ref{tab:math_prompts} for the final prompts.
}\label{tab:math_results}
\end{table}

%% file: tables/mistral_results.tex
\begin{table}[t]
\centering
\setlength{\tabcolsep}{0.45em} 
\scalebox{0.72}{
\begin{tabular}{l|cc|cccc}
\toprule
\textbf{Method}         & \textbf{GSM8k} & \textbf{MultiA.} & \textbf{Date} & \textbf{Hyper.} & \textbf{Temp.} & \textbf{Word} \\\midrule
Init. & 48.1  & 71.5    & 36   & 52     & 50    & 4    \\
Iter. APE      & 49.7  & 73.5    & 48   & 48     & 42    & 20   \\
APO            & 51.0  & 73.5    & 48   & 72     & 52    & 16   \\
PE2            & 50.5  & 74.3    & 56   & 74     & 62    & 28  \\\bottomrule
\multicolumn{7}{l}{}\\[-1em]
\multicolumn{7}{l}{\small \textbf{Tasks from BIG-Bench Hard:} Date = Date Understanding; }\\
\multicolumn{7}{l}{\small Hyper. = Hyperbaton; Temp. = Temporal Sequence, Word = Word Sorting.}
\end{tabular}
}
\vspace{-0.2cm}
\caption{Results on six selected tasks when using \texttt{Mistral-7B-Instruct-v0.2} \cite{jiang2023mistral} as the task model and \texttt{gpt-4-turbo} as the prompt proposal model. See Table~\ref{table:mistral-prompts} for the final prompts.}\label{table:mistral_results}
\end{table}

%% file: tables/math_prompts.tex
\begin{table}[t]
\centerline{
\scalebox{0.72}{
\begin{tabular}{l|p{0.5\textwidth}}
\toprule
\textbf{Method}  & \textbf{MultiArith Prompt}   \\\midrule
\rowcolor{gray!20}\multicolumn{2}{l}{Fixed Prompt}    \\\midrule
Zero-shot CoT & Let's think step by step.    \\\midrule
\multirow{2}{*}{APE}& Let’s work this out in a step by step way to be sure we have the right answer.  \\\midrule
\rowcolor{gray!20}\multicolumn{2}{l}{Prompt Optimization}  \\\midrule
Iterative APE  & Let's proceed in a methodical, step-by-step manner. \\\midrule
\multirow{3}{*}{APO} & Given the scenario, perform the necessary calculations step by step to find the final result. Consider all parts of the input and the sequence of events. \\\midrule
\multirow{3}{*}{\begin{tabular}[c]{@{}l@{}}PE2\\ (this work)\end{tabular}}  & Let's solve this problem by considering all the details. Pay attention to each piece of information, remember to add or subtract as needed, and perform the calculations step by step.  \\\bottomrule      
\end{tabular}
}}
\caption{MultiArith prompts found by compared prompt optimization methods. We use \texttt{text-davinci-003} as the task model and \texttt{gpt-4} as the prompt proposal model.}\label{tab:math_prompts}
\end{table}

%% file: tables/ablation.tex
\begin{table}[t]
\centering
\scalebox{0.72}{
\begin{tabular}{l|cc}
\toprule
\textbf{Method} & \begin{tabular}[c]{@{}c@{}}\textbf{MultiArith}\\ \textbf{Dev}\end{tabular} & \begin{tabular}[c]{@{}c@{}}\textbf{GSM8K}\\ \textbf{Dev}\end{tabular} \\\midrule
PE2 (default)           & 92.0 & 68.0 \\\midrule
\rowcolor{gray!20}\multicolumn{3}{l}{Baselines} \\\midrule
Iterative APE & 89.0 & 66.0 \\
APO  & 86.0 & 60.0 \\\midrule
\rowcolor{gray!20}\multicolumn{3}{l}{Ablation: Meta-prompt Components} \\\midrule
- two-step task description      & 89.0 & 66.0 \\
- step-by-step reasoning template    & 87.0 & 61.0 \\
- context specification & 93.0 & 63.0 \\\midrule
\rowcolor{gray!20}\multicolumn{3}{l}{Ablation: Search Algorithm Configurations} \\\midrule
- back-tracking         & 90.0 & 66.0 \\
- hard negative sampling & 90.0 & 68.0 \\\midrule
\end{tabular}
}
\caption{Ablation study on meta-prompt components. 
}\label{tab:ablation}
\end{table}
\begin{figure}
\includegraphics[width=0.48\textwidth, clip, trim=0 0 0 1.1em]{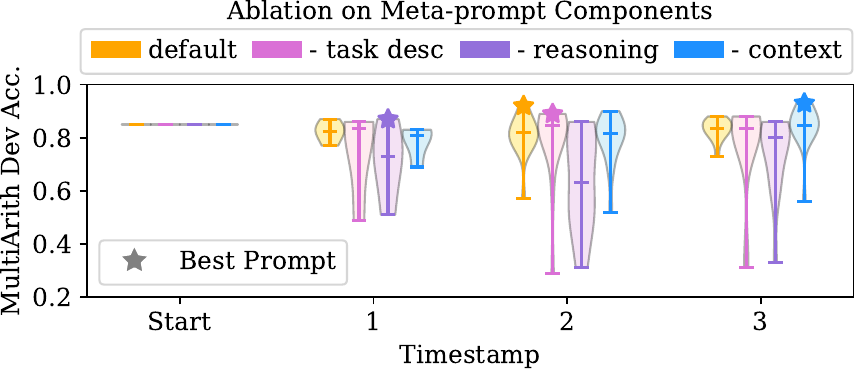}
\vspace{-0.5cm}
\caption{Prompt optimization dynamics on MultiArith when removing meta-prompt components. By removing one component, the new prompts have larger variance in their quality.
}
\label{fig:ablation}
\end{figure}

%% file: tables/successful_edits.tex
\begin{table*}[t]
\centering

\scalebox{0.72}{
\begin{tabular}{llp{0.9\textwidth}c}
\toprule
\textbf{Task}                & $t$ & \textbf{Prompt}     & \textbf{Dev Acc.} \\\midrule
\rowcolor{green!15}\multicolumn{4}{l}{Correct wrong or incomplete task instructions}  \\\midrule
\multirow{3}{*}{Rhymes}              & 0    & Remove the first letter from each input word and then {\color{Red} replace that first letter with a similar sounding letter or group of letters} to form a new word.                     & 0.35            \\
                    & 1    & Generate a word that {\color{Green} rhymes with the input word}.                                                                                                                           & 0.45            \\\midrule
\rowcolor{green!15}\multicolumn{4}{l}{Lay out tailored multi-step plans for complex problems}  \\\midrule
\multirow{6}{*}{\begin{tabular}[c]{@{}l@{}}Movie\\Recommendation\end{tabular}}  & 0    & Let's think step by step.                                                                                                                                       & 0.58            \\
                    & 1    & Consider the {\color{Green} genre, plot, and style} of the input movies. Using this information, think step by step to identify which of the following options is most similar to the given movies.                                                                                                                       & 0.74            \\
                    & 2    & Considering factors such as genre, {\color{Green} director, actors, release period, audience target, animation style, and humor}, analyze the similarities among the given movies and identify the movie from the options that shares the most similarities.                                                                                                      & 0.82           \\\midrule
\multirow{3}{*}{\begin{tabular}[c]{@{}l@{}}Date\\Understanding\end{tabular}}  & 0    & Let's think step by step.                                                                                                                                       & 0.39            \\
                    & 2    & Analyzing the given information, let's calculate the solution. Remember to consider the context provided, such as  {\color{Green}references to 'today' or specific dates}.                                                                                                     & 0.54           \\\midrule
\rowcolor{yellow!20}\multicolumn{4}{l}{Produce shortcut solutions in counterfactual tasks}  \\\midrule
\multirow{4}{*}{\begin{tabular}[c]{@{}l@{}}Base-8 Addition\\ (Induction Init.)\end{tabular}} & 0 & Add the two numbers given as input to get the output. & 0.0 \\
& 3 & Add the two numbers provided in the input. Then, adjust this sum based on the following rule: {\color{Tan}if both numbers are less than 50, add 2 to the sum. If either number is 50 or greater, add 22 to the sum.} The final result is the output. & 0.35 \\
                    \bottomrule
\end{tabular}
}
\caption{Notable prompt edits made by PE2. See \S\ref{ssec:pe2_behavior} for discussion. See Table~\ref{table:case_study2} for additional examples.
}\label{table:case_study}
\end{table*}

%% file: tables/failure_edits.tex
\begin{table*}[t]
\centering
\scalebox{0.72}{
\begin{tabular}{lp{1.1\textwidth}}
\toprule
\textbf{Task}                & \textbf{Meta-prompt and Reasoning Snippets}  \\\midrule
\rowcolor{red!15}\multicolumn{2}{l}{Neglecting instructions in the meta-prompt}  \\\midrule
\multirow{4}{*}{Base-9 Addition} & \textbf{Meta-prompt:} ... Note that {\color{Green} the ground-truth labels are \_\_absolutely correct\_\_}, but the prompts (task descriptions) may be incorrect and need modification. ...   \\\cmidrule{2-2}
 & \textbf{Reasoning:} No, it is not necessary to edit the prompt. The prompt is correct, but {\color{Red} the label is incorrect}. ... {\color{Red}The issue seems to be with the label}, not the prompt. \\\midrule
\rowcolor{red!15}\multicolumn{2}{l}{Hallucination (when hints are provided in the meta-prompt)}  \\\midrule
\multirow{6}{*}{Base-8 Addition}  & \textbf{Hint:} The calculation may be performed in a different numerical base.\\\cmidrule{2-2}
 & \textbf{Reasoning:} ... Given this, it's possible that the numbers are being added {\color{Red}in base 80}, not base 10. {\color{Red}In base 80, adding 20 to the sum of two numbers would be equivalent to adding 1 in base 10.} \\\cmidrule{2-2}
 & \textbf{New Prompt:} The inputs are two numbers separated by a '+'. {\color{Red}Add these two numbers together in base 80}, then add 1 to give the output in base 10. \\\bottomrule  
\end{tabular}
}
\caption{Limitations and failure cases of PE2. See \S\ref{ssec:pe2_behavior} for discussion.}\label{table:limitation}
\end{table*}

%% file: tables/initialization.tex
\begin{table}[ht]
\centering
\scalebox{0.7}{
\begin{tabular}{l|cc}
\toprule
\textbf{Initialization} & \begin{tabular}[c]{@{}c@{}}\textbf{MultiArith}\\ \textbf{Dev}\end{tabular} & \begin{tabular}[c]{@{}c@{}}\textbf{GSM8K}\\ \textbf{Dev}\end{tabular} \\\midrule
default (Let's think step by step.)  & 92.0 & 68.0 \\\midrule
misleading$^\dagger$ (Don't think. Just feel.) & 81.0 & 50.0 \\
irrelevant$^\dagger$ (It's a beautiful day.)& 73.0 & 49.0\\
induction from few-shot examples & 84.0 & 43.0 \\
\midrule
\rowcolor{gray!15} no-op (Let's think step by step.) & 85.0 & 57.0 \\
\bottomrule
\end{tabular}
}
\caption{Effect of Initialization. $^\dagger$ The prompts are originally from \cite{kojima2022large}.}\label{tab:initialization}
\end{table}

%% file: tables/task_format.tex
\begin{table}[ht]
\centerline{
\scalebox{0.64}{
\begin{tabular}{l|p{0.45\textwidth}}
\toprule
\textbf{Task}  & \textbf{Example}   \\\midrule
\multirow{2}{*}{\begin{tabular}[c]{@{}l@{}}Date Understanding\\ (multi-choice)\end{tabular}} & Today is 9/7. Jane is watching NFL 2003. What is the date tomorrow in MM/DD/YYYY?
Options:
(A) 09/08/1916
(B) 09/13/2003
(C) 08/18/2003
(D) 09/08/2003
(E) 09/15/2003
(F) 09/01/2003 \textbf{\textcolor{blue}{(D)}}   \\\midrule
\multirow{2}{*}{\begin{tabular}[c]{@{}l@{}}Date Understanding\\ (generative)\end{tabular}} & May 6, 1992 is like yesterday to Jane, but that is actually ten years ago. What is the date a month ago in MM/DD/YYYY? \textbf{\textcolor{blue}{04/06/2002}}  \\\midrule
\multirow{2}{*}{\begin{tabular}[c]{@{}l@{}}Movie Recommendation\\ (multi-choice 1)\end{tabular}} & Find a movie similar to Rocky, Star Wars Episode IV - A New Hope, Toy Story, The Terminator:
Options:
(A) Dracula Dead and Loving It
(B) Independence Day
(C) The Extraordinary Adventures of Adèle Blanc-Sec
(D) The American President \textbf{\textcolor{blue}{(B)}}   \\\midrule
\multirow{2}{*}{\begin{tabular}[c]{@{}l@{}}Movie Recommendation\\ (multi-choice 2)\end{tabular}} & What movie is simlar to Apollo 13, Jurassic Park, Die Hard With a Vengeance, Forrest Gump? Choose from the following: Killer Movie, Stealth, The Last Man on Earth, True Lies. \textbf{\textcolor{blue}{True Lies}} \\\midrule
\end{tabular}
}}

\caption{Different Task Formats for Date Understanding and Movie Recommendation. The correct answer is marked in \textbf{\textcolor{blue}{blue}}.}\label{tab:task_format}
\end{table}

%% file: tables/gsm8k_prompts.tex
\begin{table}[ht]
\centerline{
\scalebox{0.64}{
\begin{tabular}{l|p{0.55\textwidth}}
\toprule
\textbf{Method}  & \textbf{GSM8K Prompt}   \\\midrule
Zero-shot CoT & Let's think step by step.    \\\midrule
\multirow{3}{*}{APO} & Given the scenario, perform necessary calculations and provide a step-by-step explanation to arrive at the correct numerical answer. Consider all information provided. \\\midrule
\multirow{2}{*}{\begin{tabular}[c]{@{}l@{}}PE2\end{tabular}}  & Let's solve the problem step-by-step and calculate the required total value correctly.\\\midrule
\multirow{1}{*}{OPRO (1)} & Take a deep breath and work on this problem step-by-step.  \\\midrule
\multirow{2}{*}{OPRO (2)} & Let's combine our numerical command and clear thinking to quickly and accurately decipher the answer. \\\bottomrule      
\end{tabular}
}}

\caption{Prompts used in transferability study in Fig.~\ref{fig:transferability}.}\label{tab:gsm8k_prompts}
\end{table}

%% file: tables/ablation2.tex
\begin{table}
\centering

\scalebox{0.7}{
\begin{tabular}{l|cc}
\toprule
\textbf{Method} & \begin{tabular}[c]{@{}c@{}}\textbf{MultiArith}\\ \textbf{Dev}\end{tabular} & \begin{tabular}[c]{@{}c@{}}\textbf{GSM8K}\\ \textbf{Dev}\end{tabular} \\\midrule
PE2 (default)           & 92.0 & 68.0 \\\midrule
\rowcolor{gray!20}\multicolumn{3}{l}{Meta-prompt Components} \\\midrule
+ prompt engineering tutorial & 90.0 & 63.0 \\
+ tune batch size $\{1,2,4,8 \}$ & 92.0 & 68.0 \\
+ tune step size $\{5,10,15, \text{None}\}$ & 95.0 & 68.0 \\
+ optim history and momentum & 93.0 & 67.0 \\\bottomrule
\end{tabular}
}
\caption{Investigation on meta-prompt components and configurations (Appendix). 
}\label{table:ablation2}
\end{table}

%% file: tables/ii.tex
\begin{table*}[t]
\centerline{
\scalebox{0.74}{
\begin{tabular}{lp{0.5\linewidth}p{0.5\linewidth}}
\toprule
\textbf{Task}  & \textbf{Instruction} & \textbf{Demonstration} \\\midrule
\rowcolor{gray!20}\multicolumn{3}{l}{Subtasks used in this work (14)}                                  \\\midrule
Second Letter & Extract the first letter of the input word. & cat $\rightarrow$ a \\
Starting With & Extract the words starting with a given letter from the input sentence. & The man whose car I hit last week sued me. [m] $\rightarrow$ man, me \\
Negation & Negate the input sentence. & Time is finite $\rightarrow$ Time is not finite. \\
Antonyms & Write a word that means the opposite of the input word. & won $\rightarrow$ lost \\
Synonyms & Write a word with a similar meaning to the input word. & alleged $\rightarrow$ supposed \\
Membership & Write all the animals that appear in the given list. & cat, helicopter, cook, whale, frog, lion $\rightarrow$ frog, cat, lion, whale \\
Rhymes & Write a word that rhymes with the input word. & sing $\rightarrow$ ring \\
Informal to Formal & Rephrase the sentence in formal language. & Please call once you get there $\rightarrow$ Please call upon your arrival. \\
Translation EN-DE & Translate the word into German. & game $\rightarrow$ spiel\\
Translation EN-ES & Translate the word into Spanish. & game $\rightarrow$ jeugo\\
Translation EN-FR & Translate the word into French. & game $\rightarrow$ jeu\\
Sentiment & Determine whether a movie review is positive or negative. & The film is small in scope, yet perfectly formed. $\rightarrow$ positive\\
Sentence Similarity & Rate the semantic similarity of two sentences on a scale of 0 to 5 & Sentence 1: A man is smoking. Sentence 2: A man is skating. $\rightarrow$ 0 - definitely not\\
Word in Context & Determine whether an input word has the same meaning in the two sentences.& Sentence 1: Approach a task. Sentence
2: To approach the city. Word: approach $\rightarrow$ not the same\\ \midrule
\rowcolor{gray!20}\multicolumn{3}{l}{Subtasks removed due to near-perfect accuracy (95\%+) with baseline method (8)}      \\\midrule
First Letter & Extract the first letter of the input word. & cat $\rightarrow$ c \\
List Letters & Break the input word into letters, separated by spaces. & cat $\rightarrow$ c a t\\
Singular to Plural & Convert the input word to its plural form. & cat $\rightarrow$ cats \\
Active to Passive & Write the input sentence in passive form. & The artist introduced the scientist. $\rightarrow$ The scientist was introduced by the artist. \\
Larger Animal & Write the larger of the two given animals. & koala, snail $\rightarrow$ koala \\
Sum & Sum the two given numbers. & 22 10 $\rightarrow$ 32\\
Diff & Subtract the second number from the first. & 32 22 $\rightarrow$ 10\\
Number to Word & Write the number in English words. & 26 $\rightarrow$ twenty-six \\\midrule
\rowcolor{gray!20}\multicolumn{3}{l}{Subtask removed due to small dataset size (2)}      \\\midrule
Cause and Effect & Find which of the two given cause and effect sentences is the cause. & Sentence 1: The soda went flat. Sentence 2: The bottle was left open. $\rightarrow$ The bottle was left open.\\
Common Concept & Find a common characteristic for the given objects. & guitars, pendulums, neutrinos $\rightarrow$ involve oscillations. \\\bottomrule
\end{tabular}
}
}
\caption{Details of Instruction Induction dataset. Adapted from Table 4 in \citet{honovich-etal-2023-instruction}.}\label{tab:ii}
\end{table*}

%% file: tables/cf.tex
\begin{table*}[t]
\centerline{
\scalebox{0.75}{
\begin{tabular}{lll}
\toprule
\textbf{Task}  & \textbf{Category} & \textbf{Demonstration} \\\midrule
\rowcolor{gray!20}\multicolumn{3}{l}{Arithmetic - Two-digit addition}                                  \\\midrule
Base-10 & Original  & 22+10 $\rightarrow$ 32 \\
Base-8 & Counterfactual  & 76+76 $\rightarrow$ 174 \\
Base-9 & Counterfactual  & 76+14 $\rightarrow$ 101 \\ 
Base-11 & Counterfactual  & 76+14 $\rightarrow$ 8A \\ 
Base-16 & Counterfactual  & EC+DD $\rightarrow$ 1C9 \\ \midrule
\rowcolor{gray!20}\multicolumn{3}{l}{Chess - Legality of a 4-move opening}                                  \\\midrule
Normal Rules & Original & 1. g3 Ng6 2. b3 Kf8 * $\rightarrow$ illegal \\
Swapping bishops and knights & Counterfactual & 1. g3 Ng6 2. b3 Kf8 * $\rightarrow$ legal \\\midrule
\rowcolor{gray!20}\multicolumn{3}{l}{Syntax - Identify the main subject and the main verb of a sentence}                                  \\\midrule
SVO & Original & he has good control . $\rightarrow$ he has \\ 
SOV & Counterfactual & he good control has . $\rightarrow$ he has\\
VSO & Counterfactual & has he good control . $\rightarrow$ he has\\
VOS & Counterfactual & has good control he . $\rightarrow$ he has\\
OVS & Counterfactual & good control has he . $\rightarrow$ he has\\
OSV & Counterfactual & good control he has . $\rightarrow$ he has\\\bottomrule
\end{tabular}
}
}
\caption{Details of Couterfactual Evaluation dataset \cite{wu2023reasoning}.}\label{tab:cf}
\end{table*}

%% file: tables/successful_edits2.tex
\begin{table*}[ht]
\centering
\scalebox{0.75}{
\begin{tabular}{llp{0.9\textwidth}c}
\toprule
\textbf{Task}                & $t$ & \textbf{Prompt}     & \textbf{Dev Acc.} \\\midrule
\rowcolor{green!15}\multicolumn{4}{l}{Correct wrong or incomplete task instructions}  \\\midrule
\multirow{3}{*}{Antonyms}  & 0    & Write the opposite of the given word by {\color{Red} adding an appropriate prefix}.                                                                                                      & 0.3             \\
                    & 1    & Find the opposite of the given word. {\color{Green}If applicable, add or remove an appropriate prefix} to form the opposite.                                                              & 0.6             \\\midrule
\rowcolor{green!15}\multicolumn{4}{l}{Provide more specific context and details}  \\\midrule
\multirow{3}{*}{Second Word Letter}  & 0    & Find the second letter in each word.                                                                                                                                       & 0.9             \\
                    & 1    & Identify the second character in the provided word.                                                                                                                        & 0.95            \\
                    & 2    & Identify the second character {\color{Green}from the start of} the given word.                                                                                                            & 1.0             \\\midrule
\multirow{4}{*}{Sentence Similarity} & 0    & Rate the similarity between Sentence 1 and Sentence 2 on a scale from 1 to 5, with 1 being 'probably not similar' and 5 being 'perfectly similar'.                         & 0.0               \\
                    & 1    & Rate the similarity between Sentence 1 and Sentence 2 as {\color{Green}'1 - probably not similar', '2 - possibly', '3 - moderately', '4 - almost perfectly', or '5 - perfectly similar'}. & 0.15   \\\midrule
\rowcolor{green!15}\multicolumn{4}{l}{Lay out tailored multi-step plans for complex problems}  \\\midrule
\multirow{3}{*}{\begin{tabular}[c]{@{}l@{}}Date\\Understanding\end{tabular}}  & 0    & Let's think step by step.                                                                                                                                       & 0.39            \\
                    & 2    & Analyzing the given information, let's calculate the solution. Remember to consider the context provided, such as  {\color{Green}references to 'today' or specific dates}.                                                                                                     & 0.54           \\\midrule
\rowcolor{yellow!20}\multicolumn{4}{l}{Produce short-cut solutions in counterfactual tasks}  \\\midrule
\multirow{2}{*}{\begin{tabular}[c]{@{}l@{}}Base-9 Addition\\ (Induction Init.)\end{tabular}} & 0 & Add the numbers in each input together to get the output. & 0.0 \\
& 1 & Add the numbers in each input together and {\color{Tan}then add 11 to get the output}. & 0.2 \\
                    \bottomrule
\end{tabular}
}
\caption{Notable prompt edits made by PE2 (Part 2; Continued from Table~\ref{table:case_study}).
}\label{table:case_study2}
\end{table*}

%% file: tables/raw_results_ii.tex
\begin{table*}[ht]
\centering
\scalebox{0.75}{
\begin{tabular}{lcccccccc}
\toprule
                          & \multicolumn{2}{c}{APE}                            & \multicolumn{2}{c}{Iter. APE}                      & \multicolumn{2}{c}{APO}                            & \multicolumn{2}{c}{PE2}                                                                                        \\
                          & \multicolumn{1}{c}{mean} & \multicolumn{1}{c}{std} & \multicolumn{1}{c}{mean} & \multicolumn{1}{c}{std} & \multicolumn{1}{c}{mean} & \multicolumn{1}{c}{std} & \multicolumn{1}{c}{mean}                               & \multicolumn{1}{c}{std}                               \\\midrule
antonyms                  & 77.60                  & 3.01                  & 77.00                  & 3.63                  & 77.00                  & 2.97                  & 78.80                                                & 3.97                                                \\
informal\_to\_formal      & 59.53                  & 3.37                  & 48.83                  & 5.83                  & 54.10                  & 10.61                 & \cellcolor[HTML]{FFFFFF}{\color[HTML]{000000} 61.26} & \cellcolor[HTML]{FFFFFF}{\color[HTML]{000000} 4.73} \\
negation                  & 77.80                  & 2.48                  & 78.20                  & 2.79                  & 75.40                  & 7.17                  & 76.00                                                & 7.24                                                \\
orthography\_starts\_with & 63.80                  & 2.14                  & 65.40                  & 2.06                  & 68.60                  & 2.50                  & 67.60                                                & 1.74                                                \\
rhymes                    & 25.60                  & 12.52                 & 34.00                  & 24.26                 & 56.75                  & 22.72                 & 65.00                                                & 19.88                                               \\
second\_word\_letter      & 76.20                  & 15.12                 & 80.40                  & 15.23                 & 94.20                  & 2.32                  & 94.20                                                & 1.17                                                \\
sentence\_similarity      & 18.40                  & 4.13                  & 16.20                  & 3.19                  & 22.20                  & 15.14                 & 20.00                                                & 9.84                                                \\
sentiment                 & 88.20                  & 2.79                  & 88.20                  & 2.79                  & 88.80                  & 2.79                  & 88.80                                                & 2.79                                                \\
synonyms                  & 10.40                  & 5.20                  & 15.40                  & 1.74                  & 27.60                  & 11.71                 & 27.80                                                & 8.84                                                \\
taxonomy\_animal          & 80.80                  & 9.06                  & 83.40                  & 6.77                  & 88.80                  & 7.86                  & 89.00                                                & 8.76                                                \\
translation\_en-de        & 85.00                  & 0.89                  & 85.00                  & 0.89                  & 84.60                  & 0.80                  & 84.40                                                & 0.80                                                \\
translation\_en-es        & 84.80                  & 0.98                  & 85.00                  & 0.89                  & 85.40                  & 0.80                  & 85.40                                                & 0.49                                                \\
translation\_en-fr        & 71.80                  & 9.99                  & 75.60                  & 3.72                  & 81.80                  & 3.66                  & 80.00                                                & 3.22                                                \\
word\_in\_context         & 56.40                  & 6.92                  & 56.60                  & 7.09                  & 58.60                  & 7.66                  & 61.00                                                & 1.67                                                \\
\midrule
14-task average           & 62.60  & -    & 63.52    &-   & 68.85    & -        & 69.95  & -                  \\
($\Delta$ with APE) & (+0.00) & - & (+0.92) & - & (+6.25) & - & (+7.35) & - \\\bottomrule
\end{tabular}
}
\caption{Raw Results on Instruction Induction Benchmark. We report mean and standard deviation across 5 runs (5 different random samples of train and dev sets). The results for each task are visualized in Fig.~\ref{fig:ii} and the average results for 14 tasks are visualized in Fig.~\ref{fig:results_overview}.}\label{table:ii_raw}
\end{table*}

%% file: tables/raw_results_counterfactual.tex
\begin{table*}[]
\centering
\scalebox{0.75}{
\begin{tabular}{lcccccccc}
\toprule
                   & \multicolumn{2}{c}{APE}                            & \multicolumn{2}{c}{Iter. APE}                      & \multicolumn{2}{c}{APO}                               & \multicolumn{2}{c}{PE2}                               \\
                   & mean                        & std                  & mean                        & std                  & \multicolumn{1}{c}{mean}    & \multicolumn{1}{c}{std} & \multicolumn{1}{c}{mean}    & \multicolumn{1}{c}{std} \\\midrule
arithmetic\_base8  & 19.00                     & 15.53              & 20.20                     & 16.57              & 38.80                     & 2.56                  & 37.80                     & 3.54                  \\
arithmetic\_base9  & 11.20                     & 15.38              & 12.00                     & 16.84              & 23.80                     & 19.74                 & 29.40                     & 15.32                 \\
arithmetic\_base11 & 2.40                      & 1.50               & 2.20                      & 1.33               & 5.00                      & 2.19                  & 10.40                     & 5.50                  \\
arithmetic\_base16 & 34.40                     & 10.61              & 30.60                     & 6.09               & 55.40                     & 5.46                  & 58.00                     & 3.10                  \\
chess\_original    & 57.60                     & 4.13               & 56.00                     & 3.52               & 59.20                     & 2.04                  & 61.60                     & 3.01                  \\
chess\_cf          & 40.00                     & 3.58               & 40.20                     & 3.12               & 43.00                     & 4.15                  & 49.60                     & 5.57                  \\
syntax\_svo        & 46.20                     & 8.57               & 49.80                     & 9.24               & 57.80                     & 7.88                  & 63.40                     & 7.74                  \\
syntax\_sov        & 38.20                     & 8.28               & 44.60                     & 3.14               & 46.20                     & 7.47                  & 55.60                     & 4.80                  \\
syntax\_osv        & 11.60                     & 8.89               & 13.60                     & 10.71              & 27.80                     & 11.02                 & 41.80                     & 4.79                  \\
syntax\_ovs        & 30.00                     & 10.58              & 30.00                     & 10.58              & 33.80                     & 6.01                  & 45.20                     & 5.91                  \\
syntax\_vos        & 31.60                     & 14.37              & 30.60                     & 13.92              & 28.80                     & 9.13                  & 44.00                     & 10.45                 \\
syntax\_vso        & 26.00                     & 11.35              & 24.40                     & 11.64              & 36.80                     & 11.41                 & 42.40                     & 7.42                  \\\midrule
12-task average    & 29.02 & - & 29.52 & - & 38.03 & -  & 44.93 & - \\
($\Delta$ with APE) & (+0.00) & - & (+0.50) & - & (+9.01) & - & (+15.91) & -  \\\bottomrule
\end{tabular}
}
\caption{Raw Results on Counterfactual Eval (Induction Initialization). We report mean and standard deviation across 5 runs (5 different random samples of train and dev sets). The results for each task are visualized in Fig.~\ref{fig:results-cf} and the average results for 12 tasks are visualized in Fig.~\ref{fig:results_overview}.}\label{table:cf_raw1}
\end{table*}

\begin{table*}[]
\centering
\scalebox{0.75}{
\begin{tabular}{lcccccccc}
\toprule
                   & \multicolumn{2}{c}{Initialization}                            & \multicolumn{2}{c}{Iter. APE}                      & \multicolumn{2}{c}{APO}                               & \multicolumn{2}{c}{PE2}                               \\
                   & mean                        & std                  & mean                        & std                  & \multicolumn{1}{c}{mean}    & \multicolumn{1}{c}{std} & \multicolumn{1}{c}{mean}    & \multicolumn{1}{c}{std} \\\midrule
arithmetic\_base8  & 16.80                     & 3.71                  & 21.00                     & 5.10                  & 32.00                     & 4.56                  & 28.20                     & 3.06                  \\
arithmetic\_base9  & 2.60                      & 1.20                  & 2.00                      & 2.19                  & 7.40                      & 3.01                  & 9.80                      & 2.14                  \\
arithmetic\_base11 & 4.80                      & 1.94                  & 5.40                      & 1.85                  & 5.40                      & 1.85                  & 5.20                      & 1.72                  \\
arithmetic\_base16 & 25.60                     & 2.80                  & 33.20                     & 5.27                  & 52.00                     & 7.87                  & 52.20                     & 6.18                  \\
chess\_original    & 60.20                     & 1.94                  & 60.60                     & 3.07                  & 59.00                     & 5.02                  & 59.60                     & 5.08                  \\
chess\_cf          & 46.40                     & 2.15                  & 46.80                     & 2.14                  & 47.60                     & 3.01                  & 56.40                     & 3.67                  \\
syntax\_svo        & 0.00                      & 0.00                  & 12.80                     & 16.03                 & 44.60                     & 11.84                 & 58.40                     & 7.31                  \\
syntax\_sov        & 0.00                      & 0.00                  & 12.20                     & 14.32                 & 37.00                     & 18.83                 & 46.80                     & 11.30                 \\
syntax\_osv        & 0.00                      & 0.00                  & 8.80                      & 15.12                 & 32.40                     & 16.50                 & 49.60                     & 6.18                  \\
syntax\_ovs        & 0.00                      & 0.00                  & 0.20                      & 0.40                  & 27.20                     & 17.72                 & 45.60                     & 4.18                  \\
syntax\_vos        & 0.00                      & 0.00                  & 8.00                      & 10.37                 & 18.20                     & 15.35                 & 43.20                     & 1.72                  \\
syntax\_vso        & 0.00                      & 0.00                  & 4.60                      & 8.21                  & 20.00                     & 17.98                 & 52.80                     & 7.14                  \\\midrule
12-task average    & 13.03 & - & 17.97 & - & 31.90 & -  & 42.32 & - \\
($\Delta$ with Initialization) & (+0.00) & - & (+4.94) & - & (+18.87) & - & (+29.29) & -  \\\bottomrule
\end{tabular}
}
\caption{Raw Results on Counterfactual Eval (Manual Initialization). We report mean and standard deviation across 5 runs (5 different random samples of train and dev sets). We use ``Let's think step by step'' as the manual initialization prompt. The results for each task are visualized in Fig.~\ref{fig:cf_init}.}\label{table:cf_raw2}
\end{table*}

%% file: tables/raw_results_bbh.tex
\begin{table*}[ht]
\centering
\scalebox{0.75}{
\begin{tabular}{lcccc}
\toprule
                                            & Init. & Iter. APE & APO     & PE2     \\\midrule
boolean\_expressions                        & 88.00        & 86.00   & 94.00 & 92.00 \\
causal\_judgement                           & 48.28        & 49.43   & 58.62 & 56.32 \\
date\_understanding                         & 74.00        & 72.00   & 72.00 & 76.00 \\
disambiguation\_qa                          & 46.00        & 38.00   & 60.00 & 50.00 \\
dyck\_languages                             & 12.00        & 6.00    & 4.00  & 8.00  \\
formal\_fallacies                           & 52.00        & 50.00   & 42.00 & 52.00 \\
geometric\_shapes                           & 36.00        & 44.00   & 48.00 & 40.00 \\
hyperbaton                                  & 86.00        & 92.00   & 92.00 & 90.00 \\
logical\_deduction\_five\_objects           & 56.00        & 44.00   & 58.00 & 60.00 \\
logical\_deduction\_seven\_objects          & 40.00        & 38.00   & 40.00 & 42.00 \\
logical\_deduction\_three\_objects          & 66.00        & 82.00   & 70.00 & 66.00 \\
movie\_recommendation                       & 54.00        & 66.00   & 68.00 & 70.00 \\
multistep\_arithmetic\_two                  & 74.00        & 74.00   & 74.00 & 78.00 \\
navigate                                    & 58.00        & 56.00   & 74.00 & 84.00 \\
object\_counting                            & 64.00        & 70.00   & 66.00 & 74.00 \\
penguins\_in\_a\_table                      & 69.57        & 76.09   & 71.74 & 73.91 \\
reasoning\_about\_colored\_objects          & 74.00        & 60.00   & 74.00 & 58.00 \\
ruin\_names                                 & 66.00        & 64.00   & 70.00 & 72.00 \\
salient\_translation\_error\_detection      & 40.00        & 42.00   & 34.00 & 36.00 \\
snarks                                      & 53.85        & 60.26   & 67.95 & 73.08 \\
sports\_understanding                       & 64.00        & 66.00   & 70.00 & 64.00 \\
temporal\_sequences                         & 54.00        & 52.00   & 72.00 & 82.00 \\
tracking\_shuffled\_objects\_five\_objects  & 56.00        & 58.00   & 60.00 & 64.00 \\
tracking\_shuffled\_objects\_seven\_objects & 64.00        & 58.00   & 66.00 & 66.00 \\
tracking\_shuffled\_objects\_three\_objects & 66.00        & 66.00   & 62.00 & 64.00 \\
web\_of\_lies                               & 30.00        & 46.00   & 44.00 & 46.00 \\
word\_sorting                               & 52.00        & 60.00   & 68.00 & 66.00 \\\midrule
27-task average                             & 57.17          & 58.36     & 62.23   & 63.09  \\
($\Delta$ with Initialization)   & (+0.00)         & (+1.19)    & (+5.06)   & (+5.92) \\\bottomrule 
\end{tabular}
}
\caption{Raw Results on BIG-bench Hard Tasks. The results for each task are visualized in Fig.~\ref{fig:bbh} and the average results for 27 tasks are visualized in Fig.~\ref{fig:results_overview}.}\label{table:raw_bbh}
\end{table*}

%% file: tables/prompts1.tex
\begin{longtable}{l|l|p{0.65\textwidth}}
\caption{Prompts find by prompt optimization methods on math reasoning and instruction induction tasks. For instruction induction, experiments were run with 5 random data splits; In this table we report the prompts found in one run (seed=0).}\label{tab:all_prompts_start}\\
\toprule
\textbf{Task} & \textbf{Method}  & \textbf{Prompt}   \\\midrule
\rowcolor{gray!20}\multicolumn{3}{l}{Math Reasoning}    \\\midrule
\multirow{8}{*}{MultiArith} & Zero-shot CoT & Let's think step by step.    \\\cmidrule{2-3}
& APE & Let’s work this out in a step by step way to be sure we have the right answer.  \\\cmidrule{2-3}
&Iterative APE  & Let's proceed in a methodical, step-by-step manner. \\\cmidrule{2-3}
& APO & Given the scenario, perform the necessary calculations step by step to find the final result. Consider all parts of the input and the sequence of events. \\\cmidrule{2-3}
& PE2  & Let's solve this problem by considering all the details. Pay attention to each piece of information, remember to add or subtract as needed, and perform the calculations step by step.  \\\midrule
\multirow{8}{*}{GSM8K} & Zero-shot CoT & Let's think step by step.    \\\cmidrule{2-3}
& APE & Let’s work this out in a step by step way to be sure we have the right answer.  \\\cmidrule{2-3}
&Iterative APE  & Let's dissect this and tackle it gradually, one phase at a time.\\\cmidrule{2-3}
& APO & Given the scenario, perform necessary calculations and provide a step-by-step explanation to arrive at the correct numerical answer. Consider all information provided.\\\cmidrule{2-3}
& PE2  & Let's solve the problem step-by-step and calculate the required total value correctly.  \\\midrule
\rowcolor{gray!20}\multicolumn{3}{l}{Instruction Induction}\\\midrule
\multirow{2}{*}{antonyms}& APO & Provide the opposite or a negative form of the given input word.\\\cmidrule{2-3}
& PE2 & Provide the opposite or a negative form of the given input word.\\\midrule
\multirow{5}{*}{informal\_to\_formal}& APO & Convert each sentence into a formal version, preserving the original structure, meaning, and tone. Avoid excessive formality, unnecessary changes, and maintain idiomatic expressions. Handle contractions appropriately.\\\cmidrule{2-3}
& PE2 & Please transform each sentence into a version that maintains the original meaning but is expressed in a more formal or polite manner.\\\midrule
\multirow{2}{*}{negation}& APO & Negate the statement given in the input.\\\cmidrule{2-3}
& PE2 & Negate the statement given in the input.\\\midrule
\multirow{4}{*}{orthography\_starts\_with}& APO & Identify the word or phrase in the sentence that starts with the given letter, considering the context and grammar. Include articles if they precede the word or phrase.\\\cmidrule{2-3}
& PE2 & Find the word or phrase in the sentence that starts with the given letter, and write it as the output.\\\midrule
\multirow{2}{*}{rhymes}& APO & Remove the first letter of the given word. Find a word that rhymes with the remaining part, has the same syllable count, and is not a derivative or the same as the original word.\\\cmidrule{2-3}
& PE2 & Generate a word that rhymes with the given word.\\\midrule
\multirow{2}{*}{second\_word\_letter}& APO & Identify the second character from the start in each input word and provide it as the output.\\\cmidrule{2-3}
& PE2 & Identify the second character from the start of the given word.\\\midrule
\multirow{5}{*}{sentence\_similarity}& APO & Rate the similarity between Sentence 1 and Sentence 2 using the scale: 1 - 'probably not', 2 - 'possibly', 3 - 'probably', 4 - 'likely', 5 - 'perfectly'.\\\cmidrule{2-3}
& PE2 & Rate the similarity between Sentence 1 and Sentence 2 using the scale: 1 - 'probably not', 2 - 'possibly', 3 - 'probably', 4 - 'likely', 5 - 'perfectly'.\\\midrule
\multirow{2}{*}{sentiment}& APO & Determine if the given movie review statement is positive or negative.\\\cmidrule{2-3}
& PE2 & Determine if the given movie review statement is positive or negative.\\\midrule
\multirow{2}{*}{synonyms}& APO & Provide a single word that is closely related to the given input, considering its most common usage.\\\cmidrule{2-3}
& PE2 & Identify a word that is closely connected, in meaning or context, with the provided input word.\\\midrule
\multirow{2}{*}{taxonomy\_animal}& APO & Remove all items from the list that are not animals.\\\cmidrule{2-3}
& PE2 & Remove all items from the list that are not animals.\\\midrule
\multirow{2}{*}{translation\_en-de}& APO & Translate each English word into German.\\\cmidrule{2-3}
& PE2 & Translate each English word into German.\\\midrule
\multirow{3}{*}{translation\_en-es}& APO & Provide the most commonly used Spanish translation for the given English word.\\\cmidrule{2-3}
& PE2 & Translate the given term from English to Spanish. Note that the translation may be a single word or a phrase.\\\midrule
\multirow{2}{*}{translation\_en-fr}& APO & Provide the French equivalent for the given English word.\\\cmidrule{2-3}
& PE2 & Translate the following word from English to its most common equivalent in French.\\\midrule
\multirow{4}{*}{word\_in\_context}& APO & Determine if the word provided is used in the same sense/context in both sentences. If it is, write 'same.' If not, write 'not the same.'\\\cmidrule{2-3}
& PE2 & Determine if the word provided is used in the same sense/context in both sentences. If it is, write 'same.' If not, write 'not the same.'\\
\bottomrule  
\end{longtable}

%% file: tables/prompts2.tex
\begin{longtable}{l|l|p{0.7\textwidth}}
\caption{Prompts find by prompt optimization methods on Counterfactual Eval \cite{wu2023reasoning} using induction initialization (\textit{i.e.}, the model is not informed of the counterfactual situation). Experiments were run with 5 random data splits; In this table we report the prompts found in one run (seed=0).}\\
\toprule
\textbf{Task} & \textbf{Method}  & \textbf{Prompt}   \\\midrule
\endfirsthead
\toprule
\textbf{Task} & \textbf{Method}  & \textbf{Prompt}   \\\midrule
\endhead
\multicolumn{3}{r}{\textbf{Continued on next page}} \\
\endfoot
\bottomrule
\endlastfoot 
\rowcolor{gray!20}\multicolumn{3}{l}{Counterfactual Evaluation (Induction Initialization)}    \\\midrule
\multirow{7}{*}{arithmetic\_base11}& APO & Given two numbers in hexadecimal format (0-9, A-F), convert each number to decimal. Add the two decimal numbers together. Output the sum in hexadecimal format. If the sum exceeds the range of a single hexadecimal digit (0-F), represent it appropriately in hexadecimal. For example, if the input is 'A' and 'B', the output should be '15' as 'A' is 10 and 'B' is 11 in decimal, and their sum is 21 which is '15' in hexadecimal.\\\cmidrule{2-3}
& PE2 & Convert both numbers in each pair from hexadecimal to decimal, then add them together. Output the resultant sum in hexadecimal. For instance, if the input is A4+61, convert A4 and 61 to decimal (164 and 97 respectively), add them together to get 261, and convert this back to hexadecimal to get 105.\\\midrule
\multirow{7}{*}{arithmetic\_base16}& APO & Given two hexadecimal numbers as input, add them together using base 16 arithmetic. The input hexadecimal numbers will be in uppercase and may have different number of digits. Align the numbers from right to left, similar to traditional addition, and handle any overflow or carry appropriately. Output the sum as an uppercase hexadecimal number.\\\cmidrule{2-3}
& PE2 & Add the input hexadecimal numbers together and output the sum as a hexadecimal number. For example, if the input is "44+E7", the output should be "12B", because the sum of hexadecimals 44 and E7 equals 12B in hexadecimal.\\\midrule
\multirow{8}{*}{arithmetic\_base8}& APO & Given an input string containing two numbers separated by a '+', calculate the sum of these two numbers. Then, add 20 to this sum to get the output. For example, if the input is '22+47', first add 22 and 47 to get 69, then add 20 to 69 to get the final output of 89. Similarly, if the input is '74+26', first add 74 and 26 to get 100, then add 20 to 100 to get the final output of 120. The '+' symbol should be interpreted as an addition operator, and the order of operations should be to add the two numbers first, then add 20 to the sum. The input will always be formatted correctly, with no spaces or other characters around the '+' symbol.\\\cmidrule{2-3}
& PE2 & To find the correct output, first add the two numbers given as input. Once you have the sum of these two numbers, add an additional 22 to this sum. For example, if the input is "17+65", you should first add 17 and 65 to get 82, then add 22 to 82. The correct output in this case would be 104.\\\midrule
\multirow{4}{*}{arithmetic\_base9}& APO & Add the numbers together.\\\cmidrule{2-3}
& PE2 & Add the two numbers given as input and then add 10 to the result to generate the output. For example, if the input is '25+18', the output should be '53' because 25 plus 18 equals 43, and adding 10 gives 53.\\\midrule
\multirow{8}{*}{chess\_cf}& APO & Determine if the given sequence of chess moves, starting from the initial game position, is legal or not according to the standard rules of chess. Consider the unique movements and restrictions of each piece, the alternating turns of the players (white and black), and the entire game state up to the given point. Evaluate the sequence as a whole, not just individual moves. Note that the sequence ends with an asterisk (*).\\\cmidrule{2-3}
& PE2 & Please assess the legality of the following sequence of chess moves based on standard chess rules. If all moves are valid according to the rules of chess, indicate "Legal." If there is any move that violates standard chess rules, respond with "Illegal". For example, if the sequence is "1. e4 e5 2. Nf3 d6", your response should be "Legal". If the sequence is "1. e4 e5 2. Kf2", your response should be "Illegal" because the king cannot be exposed to check.\\\midrule
\multirow{2}{*}{chess\_original}& APO & Determine if the given sequence of chess moves is legal or illegal.\\\cmidrule{2-3}
& PE2 & Determine if the given sequence of chess moves is legal or illegal.\\\midrule
\multirow{4}{*}{syntax\_osv}& APO & Identify the main subject and verb in the sentence. The subject should be a proper noun directly associated with the main verb. Focus on the main clause that conveys the primary information. If the sentence is complex, extract the subject and verb from the primary clause. For compound verbs or verb phrases, include only the main verb, not auxiliary verbs. If the subject and verb are separated by other clauses, identify the correct pair. If the subject is implied, make a reasonable guess. Write the subject and verb as a pair in the output.\\\cmidrule{2-3}
& PE2 & Identify the subject and verb at the end of the sentence. The subject may not always be a proper noun. The verb should be in the present tense. Write them out as a pair in the output. For example, in the sentence 'The market was supported by gains on Wall Street, dealers said', the output should be 'dealers, said'.\\\midrule
\multirow{8}{*}{syntax\_ovs}& APO & Identify the first instance of a subject in the sentence, which could be a pronoun ('he', 'she', 'it', 'they', 'we', etc.) or a noun/noun phrase. Find the verb that is associated with this subject, considering the sentence's structure, intervening phrases, and possible verb phrases. The verb may not directly follow the subject and could precede it. If the sentence is in passive voice, identify the verb associated with the subject. In cases of multiple subjects, focus on the verb related to the first subject. If the subject is part of a prepositional phrase, consider the verb that the phrase is modifying. Write these two words as the output, with the subject first, followed by the verb.\\\cmidrule{2-3}
& PE2 & Identify the first personal pronoun in the sentence and find the verb that is semantically linked to it. Write these two words as your output. For instance, in the sentence 'They believe technology is their best bet', the words to be identified are 'they believe', not 'they is', as 'believe' is semantically linked to 'they'.\\\midrule
\multirow{8}{*}{syntax\_sov}& APO & Identify the main subject and the main verb in the sentence. Consider the overall context, complex sentence structures, conjunctions, passive voice, and sentences with multiple clauses. Output the main subject and the main verb together, as they appear in the input. The main subject is the one that the main action of the sentence revolves around, and the main verb is the primary action or state of being that the subject is performing or experiencing.\\\cmidrule{2-3}
& PE2 & Identify the subject and the main verb in the sentence and write them together in the same order as they appear in the sentence, excluding any additional words in between. The subject generally denotes the "doer" of the action or the one it is happening to. The main verb expresses the action or state of being. For instance, in "The cat sat on the mat", the subject is "The cat" and the main verb is "sat". So, the output should be "The cat sat". Ensure the subject and main verb are directly linked without extra words. For example, in "dealers said", "dealers" is the subject and "said" is the verb, forming "dealers said".\\\midrule
\multirow{8}{*}{syntax\_svo}& APO & Your task is to identify the subject and the main verb of the primary clause in an input sentence. Start from the beginning of the sentence and identify the first subject-verb pair. Ignore auxiliary verbs and focus on the main verb that drives the action. If the sentence has multiple clauses, focus on the first one that forms a complete thought. Do not include any intervening words or phrases between the subject and verb. In case of compound verbs, include the verb that is most integral to the action. Ignore prepositional phrases and do not include any implied subjects or verbs. Your output should be concise, containing only the subject and the main verb.\\\cmidrule{2-3}
& PE2 & Read the input sentence and identify the subject and the verb of the main clause. Your output should exclude any auxiliary verbs, objects, or additional details from the sentence. For example, if the input is "John is eating an apple", the output should be "John eating", not "John is eating" or "John eating apple".\\\midrule
\multirow{8}{*}{syntax\_vos}& APO & Identify the first and last words of each sentence, considering a sentence as a group of words that starts with a capital letter and ends with a period, question mark, or exclamation point. Ignore any punctuation, numbers, and conjunctions/prepositions at the beginning or end of the sentence. Write these two words in reverse order. If the sentence begins and ends with the same word, write it once. Treat compound words or phrases as single words. For example, 'uniroyal' and 'has' should be treated as 'uniroyal has'.\\\cmidrule{2-3}
& PE2 & Identify the main subject and verb in each input sentence and form a pair. The subject is usually a noun or pronoun that the verb refers to. The verb should be the main verb of the sentence, not an auxiliary verb. For example, if the input is "The cat chased the mouse.", the output should be "cat chased". If the input is "She has eaten the cake.", the output should be "She eaten", not "She has".\\\midrule
\multirow{8}{*}{syntax\_vso}& APO & Identify the main subject and the primary verb in the given sentence, regardless of their position or the complexity of the sentence. Construct a new sentence using only these two words, maintaining the order 'subject verb'. Ignore additional information, context, or implied subjects/verbs. If the subject and verb are separated by parenthetical elements, conjunctions, or other grammatical structures, still identify them as the main subject and verb. Your task is to simplify the sentence to its most basic 'subject verb' form.\\\cmidrule{2-3}
& PE2 & Identify the main subject and the corresponding verb in the given sentence and construct a new short sentence using only these two words. The order should be 'subject verb'. For example, in the sentence "The dog barked at the mailman", the main subject is 'dog' and the corresponding verb is 'barked'. So, the new sentence would be "Dog barked".\\
\end{longtable}

\begin{longtable}{l|l|p{0.7\textwidth}}
\caption{Prompts find by prompt optimization methods on Counterfactual Eval \cite{wu2023reasoning} using manual initialization. Experiments were run with 5 random data splits; In this table we report the prompts found in one run (seed=0).}\\
\toprule
\textbf{Task} & \textbf{Method}  & \textbf{Prompt}   \\\midrule
\endfirsthead
\toprule
\textbf{Task} & \textbf{Method}  & \textbf{Prompt}   \\\midrule
\endhead
\multicolumn{3}{r}{\textbf{Continued on next page}} \\
\endfoot
\bottomrule
\endlastfoot 
\rowcolor{gray!20}\multicolumn{3}{l}{Counterfactual Evaluation (Manual Initialization)}    \\\midrule
\multirow{8}{*}{arithmetic\_base11}& Manual Init. &You are a mathematician. Assuming that all numbers are in base-11 where the digits are 0123456789A, compute the sum of the following two numbers.\\\cmidrule{2-3}
& APO & You are a mathematician. Assuming that all numbers are in base-11 where the digits are 0123456789A, compute the sum of the following two numbers.\\\cmidrule{2-3}
& PE2 & You are a mathematician. Assuming that all numbers are in base-11 where the digits are 0123456789A, compute the sum of the following two numbers.\\\midrule
\multirow{12}{*}{arithmetic\_base16}& Manual Init. &You are a mathematician. Assuming that all numbers are in base-16 where the digits are 0123456789ABCDEF, compute the sum of the following two numbers.\\\cmidrule{2-3}
& APO & You are a mathematician working with base-16 (hexadecimal) numbers. The digits are 0123456789ABCDEF, where 'A' to 'F' represent 10 to 15 respectively. Add the two given hexadecimal numbers. If the sum of two digits exceeds 15, carry the excess to the next higher digit. For instance, 'F' + '2' equals '11' in base-16, which is '1' with a carryover of '1'. The input will be two hexadecimal numbers separated by a '+'. The output should be the sum in base-16."\\\cmidrule{2-3}
& PE2 & As a base-16 mathematician, your task is to add the provided hexadecimal numbers together. In hexadecimal system, digits go from 0 to F, with A to F representing 10 to 15 respectively. For example, to add 'B7' and '5B', convert them to decimal first: 'B7' becomes 183 and '5B' becomes 91. Their sum, 274, is '112' in hexadecimal.\\\midrule
\multirow{12}{*}{arithmetic\_base8}& Manual Init. & You are a mathematician. Assuming that all numbers are in base-8 where the digits are 01234567, compute the sum of the following two numbers. \\\cmidrule{2-3}
& APO & You are a mathematician specializing in the octal (base-8) number system. Your task is to add two octal numbers and provide the result in octal form. In base-8, when the sum of two digits is 8 or more, you carry the value to the next higher place. For example, 7+1 in base-8 is 10. Here are some examples:\\\cmidrule{2-3}
& PE2 & As a mathematician, your task is to add the following two numbers which are represented in base-8 (octal) format. The base-8 system uses digits from 0 to 7. Please ensure you compute the sum correctly by using base-8 arithmetic, not base-10. For example, in base-8, 7+1 equals 10, not 8. Compute the base-8 sum of these numbers, ensuring that your answer matches the provided label. For instance, if the input is "25+55", the correct output would be "102". Now, compute the base-8 sum of these numbers:\\\midrule
\multirow{12}{*}{arithmetic\_base9}& Manual Init. &You are a mathematician. Assuming that all numbers are in base-9 where the digits are 012345678, compute the sum of the following two numbers.\\\cmidrule{2-3}
& APO & You are a mathematician working with base-9 numbers, where digits range from 0 to 8. Your task is to add two base-9 numbers. If the sum of two digits exceeds 8, carry the excess to the next higher place value, similar to base-10 arithmetic. For instance, '8+1' in base-9 equals '10'. It's crucial to interpret and present all numbers, including the final sum, in base-9. For example, if you're adding '16' and '24' in base-9, the correct sum is '41', not '40'. Now, compute the sum of the following two base-9 numbers.\\\cmidrule{2-3}
& PE2 & You are a mathematician. Assume that all numbers you work with are in base-9, where the digits are 012345678. Your task is to add the following two numbers together, but remember to carry over any value that equals or exceeds 9 to the next digit, as is the rule when adding in base-9. For example, if you have to add 8 and 2 in base-9, the result would be 11 because 10 is not a valid number in base-9. Now, compute the sum of the following two numbers.\\\midrule
\multirow{8}{*}{chess\_cf}& Manual Init. &You are a chess player. You are playing a chess variant where the starting positions for knights and bishops are swapped. For each color, the knights are at placed that where bishops used to be and the bishops are now placed at where knights used to be. Given an opening, determine whether the opening is legal. The opening doesn't need to be a good opening. Answer "legal" if all moves are legal. Answer "illegal" if the opening violates any rules of chess.\\\cmidrule{2-3}
& APO & You are evaluating a chess variant where knights and bishops have swapped starting positions. Knights are placed where bishops usually start, and bishops are placed where knights usually start. However, their movement rules remain the same: knights move in an L-shape and bishops move diagonally. Your task is to determine the legality of a given opening. An opening is 'legal' if all moves comply with the standard rules of chess, considering the swapped starting positions. If all moves are legal, answer 'legal'. If any move violates the chess rules, answer 'illegal'. The opening doesn't need to be a good strategy, it just needs to be legal.\\\cmidrule{2-3}
& PE2 & You are a chess enthusiast, playing a variant of the game where knights and bishops have swapped their starting positions and movements. The knights, now placed where the bishops were, move as bishops. The bishops, positioned where knights were, move as knights. Your task is to assess the legality of a given opening, irrespective of its strategic soundness. Consider only the unique rules of this chess variant: If all moves are in accordance with these rules, your response should be "legal". However, if any move contravenes these rules, respond with "illegal". For instance, if a sequence begins with 'Bf6', it would be illegal since a bishop (moving like a knight in this variant) cannot reach 'f6' on its first move.\\\midrule
\multirow{8}{*}{chess\_original}& Manual Init. &You are a chess player. Given an opening, determine whether the opening is legal. The opening doesn't need to be a good opening. Answer "legal" if all moves are legal. Answer "illegal" if the opening violates any rules of chess.\\\cmidrule{2-3}
& APO & You are a chess expert. Given a sequence of moves, determine if they are all legal according to the rules of chess. Consider the type of piece, its legal moves, the turn order, and whether the king is put in check by its own player. If all moves are legal, answer "legal". If any move violates the rules of chess, answer "illegal". Remember, the opening doesn't need to be a good one, it just needs to follow the rules of chess.\\\cmidrule{2-3}
& PE2 & As a chess expert, your task is to examine the given opening sequence in a chess game and determine if it adheres to the official rules of chess. Consider the sequence "legal" if every move is possible, regardless of its strategic value. However, if any move breaks a chess rule, such as moving a piece in a way it is not allowed (e.g., a knight moving like a bishop), classify the sequence as "illegal". Your response should be one of two words: "legal" or "illegal".\\\midrule
\multirow{4}{*}{syntax\_osv}& Manual Init. &You are an expert in linguistics. Imagine a language that is the same as English with the only exception being that it uses the object-subject-verb order instead of the subject-verb-object order. Your task is to identify the main verb and the main subject in a sentence in this imaginary language. Show the main verb (a single word) and its subject (also a single word).\\\cmidrule{2-3}
& APO & You are a linguistics expert. Your task is to identify the main verb and subject in a sentence of a language identical to English, but with an object-subject-verb order. The main verb is the primary action word, excluding auxiliary verbs. The main subject is the primary entity performing the action. In complex sentences, focus on the main clause. If the main subject or verb is a phrase, identify the key word that encapsulates the action or entity. If the main subject or verb is a proper noun, treat it as a single word. Your output should be a phrase consisting of the main subject and verb. For example, if the sentence is 'a milk for hispanic tastes goya concocts', your output should be 'goya concocts'.\\\cmidrule{2-3}
& PE2 & As a linguistics expert, your task is to analyze sentences from a language that, while similar to English, employs an object-subject-verb order instead of the English subject-verb-object order. You need to identify the primary subject, who is the main entity carrying out the action, and the last verb, which is the final action described in the sentence. Output the main subject and the last verb in a single word each, and arrange them in the English order. For instance, for "apple the eats boy", your output should be "boy eats". Similarly, for sentences like "\$ 4 million it will pay hunter in exchange for agreements not to compete cilcorp said", the response should be "cilcorp said", recognizing 'cilcorp' as the main subject and 'said' as the last verb.\\\midrule
\multirow{16}{*}{syntax\_ovs}& Manual Init. &You are an expert in linguistics. Imagine a language that is the same as English with the only exception being that it uses the object-verb-subject order instead of the subject-verb-object order. Your task is to identify the main verb and the main subject in a sentence in this imaginary language. Show the main verb (a single word) and its subject (also a single word).\\\cmidrule{2-3}
& APO & You are a linguistics expert analyzing a language similar to English, but with an object-verb-subject (OVS) order. Your task is to identify the main verb and the main subject in a sentence. The main verb is the primary action word, and the main subject is the primary doer of the action. They may not always be adjacent. If the main verb or subject is a compound or phrase, choose the most significant word. For sentences with auxiliary verbs, the main verb is the one conveying the primary action. After identifying, reverse the order to subject-verb for your output. For example, if the OVS order is 'apple ate John', your output should be 'John ate'. Remember, your output should always be in subject-verb order.\\\cmidrule{2-3}
& PE2 & You are an expert in linguistics. Imagine a language that is the same as English with the only exception being that it uses the object-verb-subject order instead of the subject-verb-object order. Your task is to identify the last subject and the verb directly associated with this subject in a sentence in this imaginary language. Show the subject first (a single word) and then the verb (also a single word). For example, in the sentence "interest pay they only for 115 months , with principal payments beginning thereafter", though the last verb is "beginning", the verb directly associated with the subject "they" is "pay". Therefore, the answer is "they pay".\\\midrule
\multirow{4}{*}{syntax\_sov}& Manual Init. &You are an expert in linguistics. Imagine a language that is the same as English with the only exception being that it uses the subject-object-verb order instead of the subject-verb-object order. Your task is to identify the main verb and the main subject in a sentence in this imaginary language. Show the main verb (a single word) and its subject (also a single word).\\\cmidrule{2-3}
& APO & You are a linguistics expert. Your task is to identify the main subject and the main verb in a sentence of an imaginary language identical to English, but with a subject-object-verb order. Your output should be in the original English order (subject-verb). Choose the most crucial word if the subject or verb is a phrase. Ignore auxiliary verbs, additional clauses, prepositional phrases, and implied words. Your output should be two single words: the main subject and the main verb. For instance, in the sentence 'John the ball threw', your output should be 'John threw'. In complex sentences, focus on the primary clause. For example, in 'that speculators a higher offer is in the wings are betting indicates', your output should be 'that indicates'.\\\cmidrule{2-3}
& PE2 & As a linguistics expert, consider an alternate version of English that uses the subject-object-verb order instead of the traditional subject-verb-object order. Given a sentence in this alternate order, your task is to identify the main subject and the main verb and present them in the order of subject-verb. Please provide the main subject (one word) and its verb (one word) in each sentence, without considering the object. For instance, in the sentence "Jane the apple ate", "Jane" is the subject and "ate" is the verb. Therefore, the answer would be "Jane ate".\\\midrule
\multirow{16}{*}{syntax\_svo}& Manual Init. &You are an expert in linguistics. Your task is to identify the main verb and the main subject in a sentence. Show the main verb (a single word) and its subject (also a single word).\\\cmidrule{2-3}
& APO & You are a language analyst. Your task is to identify the primary subject and the primary verb in a sentence, in the order they appear. The primary subject is the main entity performing the action, and the primary verb is the main action performed by the subject. They should be part of the same clause. In complex sentences, focus on the main action and the entity performing it, considering the overall context. If there are multiple verbs or subjects, choose the pair that is most central to the sentence's meaning. Ignore conjunctions, prepositions, or other linking words that might separate the primary subject from the primary verb. If the primary subject or verb is implied, infer it from the context. Provide the primary subject and verb as a single output, with the subject first and the verb second. Both should be single words. Do not include punctuation in your output.\\\cmidrule{2-3}
& PE2 & As a linguistics expert, your task is to determine the main verb and the main subject in a given sentence. Identify them as a single word each. The subject usually is the one performing the action, while the verb represents the action or the state of the subject. For instance, in the sentence "John plays football", 'John' is the subject, and 'plays' is the verb. Please provide the subject first, followed by the verb.\\\midrule
\multirow{16}{*}{syntax\_vos}& Manual Init. &You are an expert in linguistics. Imagine a language that is the same as English with the only exception being that it uses the verb-object-subject order instead of the subject-verb-object order. Your task is to identify the main verb and the main subject in a sentence in this imaginary language. Show the main verb (a single word) and its subject (also a single word).\\\cmidrule{2-3}
& APO & You are a linguistics expert. Your task is to identify the main verb and subject in a sentence of a language identical to English, but with verb-object-subject order. Focus on the verb and subject that carry the main action or idea. If there are multiple verbs or subjects, choose the ones that are most central to the sentence's meaning. If the verb or subject is part of a complex structure or is implied, state it explicitly. If the verb or subject is a phrase, identify the entire phrase. Your output should be in the format: 'Subject Verb'. Remember, the subject and verb may not be adjacent or single words. Use your linguistic expertise to determine the main verb and subject.\\\cmidrule{2-3}
& PE2 & You are a linguistics expert tasked with analyzing sentences in a language similar to English but with a key difference: the order of the verb, object, and subject is changed. Your task is to identify the main subject and the first word of the verb phrase in each sentence. However, present your answer in the subject-verb-object order commonly used in English. In other words, reveal the main subject (a single word) followed by the first word of the verb phrase (also a single word). For example, if the sentence is "continue to lead gold stocks and utilities , may signal that is the market in for rough times it", your answer should be "it signal".\\\midrule
\multirow{4}{*}{syntax\_vso}& Manual Init. &You are an expert in linguistics. Imagine a language that is the same as English with the only exception being that it uses the verb-subject-object order instead of the subject-verb-object order. Your task is to identify the main verb and the main subject in a sentence in this imaginary language. Show the main verb (a single word) and its subject (also a single word).\\\cmidrule{2-3}
& APO & You are a language expert analyzing a unique language similar to English, but with verb-subject-object order. Your task is to identify the main verb and subject in a sentence. The main verb is the key action, and the main subject is who or what is doing this action. In complex sentences, focus on the most important action. If multiple verbs or subjects exist, choose the most central to the sentence's meaning. Treat auxiliary or compound verbs as one unit with their main verb. Your output should be the main subject followed by the main verb (both as single words)."\\\cmidrule{2-3}
& PE2 & As a linguistics expert, consider an alternative English language that uses verb-subject-object order instead of the standard subject-verb-object order. Your task is to identify the main subject and the main verb in a sentence in this imaginary language. Display the main subject (a single word) followed by its verb (also a single word). For instance, if the input is "compares that with 3.5 \% butterfat for whole milk", the output should be "that compares". Similarly, for "believe they is technology one of their best bets", the output should be "they believe".\\
\end{longtable}

%% file: tables/prompts3.tex
\begin{longtable}{l|l|p{0.68\textwidth}}
\caption{Prompts find by prompt optimization methods on BIG-bench Hard tasks. Here we use the task formats in \cite{suzgun-etal-2023-challenging}. In these experiments, we use \texttt{gpt-3.5-turbo-instruct} as the task model and \texttt{gpt-4} as the prompt proposal model. See Table~\ref{table:bbh} when alternative formats are used.}\\
\toprule
\textbf{Task} & \textbf{Method}  & \textbf{Prompt}   \\\midrule
\endfirsthead
\toprule
\textbf{Task} & \textbf{Method}  & \textbf{Prompt}   \\\midrule
\endhead
\multicolumn{3}{r}{\textbf{Continued on next page}} \\
\endfoot
\bottomrule
\endlastfoot 
\rowcolor{gray!20}\multicolumn{3}{l}{BIG-bench Hard (Manual Initialization - ``Let's think step by step.'')}    \\\midrule
\multirow{5}{*}{boolean\_expressions}
& Iterative APE & Consider approaching this methodically, one step at a time.\\\cmidrule{2-3}
& APO & First, resolve 'not' immediately before it. Then evaluate expressions within parentheses. Next, perform 'and' operations, and finally 'or'. Determine the truth value step by step.\\\cmidrule{2-3}
& PE2 & Evaluate the logical expression step by step following precedence: 'not', 'and', 'or'. Conclude by affirming the truth of the full statement.\\\midrule
\multirow{1}{*}{causal\_judgement}
& Iterative APE & Approach each stage sequentially.\\\cmidrule{2-3}
& APO & Identify the direct cause of the outcome: was it the immediate action or condition without which the event wouldn't have occurred?\\\cmidrule{2-3}
& PE2 & Determine if the action was intentional and a contributing factor to the outcome. Answer 'Yes' if intentional and causative, 'No' otherwise.\\\midrule
\multirow{5}{*}{disambiguation\_qa}
& Iterative APE & Consider each phase one by one.\\\cmidrule{2-3}
& APO & In each sentence, determine the pronoun's antecedent based on typical roles and actions. If unclear, mark as ambiguous."\\\cmidrule{2-3}
& PE2 & Let's think step by step, using grammatical structure and context to identify the antecedent of the pronoun or determine if it is ambiguous.\\\midrule
\multirow{5}{*}{date\_understanding}
& Iterative APE & Consider the process in a sequential manner.\\\cmidrule{2-3}
& APO & Let's think step by step.\\\cmidrule{2-3}
& PE2 & Determine the date referred to as 'today' from the context given, then calculate the specified date step by step in MM/DD/YYYY format.\\\midrule
\multirow{5}{*}{dyck\_languages}
& Iterative APE & Let's think step by step.\\\cmidrule{2-3}
& APO & Output the sequence of closing brackets needed to properly close all open brackets in the given order: parentheses '()', square '[]', curly '{}', and angle '<>'. Ignore all other characters.\\\cmidrule{2-3}
& PE2 & Given an incomplete sequence of brackets, output only the missing closing brackets required to complete the sequence. Exclude any part of the original sequence.\\\midrule
\multirow{5}{*}{formal\_fallacies}
& Iterative APE & Consider each phase one at a time.\\\cmidrule{2-3}
& APO & Determine if the conclusion necessarily follows from the premises using deductive logic. Identify any logical fallacies or overlooked counterexamples. Is the argument valid or invalid?\\\cmidrule{2-3}
& PE2 & To assess if an argument is deductively valid, examine if the conclusion logically follows from the premises without exception.\\\midrule
\multirow{5}{*}{geometric\_shapes}
& Iterative APE & Consider our actions in a sequential manner.\\\cmidrule{2-3}
& APO & Classify the shape formed by the SVG path's 'd' attribute, analyzing vertices, line segments, and closure to distinguish polygons and figures.\\\cmidrule{2-3}
& PE2 & Analyze the SVG path commands to determine the shape they draw. Consider each command and visualize the path step by step.\\\midrule
\multirow{5}{*}{hyperbaton}
& Iterative APE & Approach this gradually, step by step.\\\cmidrule{2-3}
& APO & Choose the sentence with adjectives in the correct order: opinion, size, age, shape, color, origin, material, purpose, noun."\\\cmidrule{2-3}
& PE2 & Let's think step by step, considering the standard order of adjectives in English: opinion, size, age, shape, color, origin, material, purpose.\\\midrule
\multirow{5}{*}{\begin{tabular}[c]{@{}l@{}}logical\_deduction\\\_five\_objects\end{tabular}}
& Iterative APE & Let's proceed with a methodical, step-by-step approach.\\\cmidrule{2-3}
& APO & Let's think step by step.\\\cmidrule{2-3}
& PE2 & Let's think step by step.\\\midrule
\multirow{5}{*}{\begin{tabular}[c]{@{}l@{}}logical\_deduction\\\_seven\_objects\end{tabular}}
& Iterative APE & Consider each phase one at a time.\\\cmidrule{2-3}
& APO & Arrange seven objects in order using the clues provided. Identify the correct position for each object from the options listed, ensuring logical consistency."\\\cmidrule{2-3}
& PE2 & Let's analyze the information provided step by step, ensuring each deduction follows logically from the statements given to arrive at the correct answer.\\\midrule
\multirow{5}{*}{\begin{tabular}[c]{@{}l@{}}logical\_deduction\\\_three\_objects\end{tabular}}
& Iterative APE & Let's proceed with a methodical, stepwise approach.\\\cmidrule{2-3}
& APO & Let's think step by step.\\\cmidrule{2-3}
& PE2 & Let's carefully analyze the information to determine the accurate ranking of objects based on their attributes.\\\midrule
\multirow{5}{*}{\begin{tabular}[c]{@{}l@{}}multistep\_arithmetic\\\_two\end{tabular}}
& Iterative APE & Approach this step-by-step, tackling each phase sequentially.\\\cmidrule{2-3}
& APO & Let's think step by step.\\\cmidrule{2-3}
& PE2 & Let's meticulously solve the math problem by simplifying each part of the equation, verifying our calculations at every step before moving forward.\\\midrule
\multirow{5}{*}{\begin{tabular}[c]{@{}l@{}}movie\\\_recommendation\end{tabular}}
& Iterative APE & Consider each phase methodically.\\\cmidrule{2-3}
& APO & Select the movie that best matches the reference in terms of genre and cultural significance, ignoring minor details. Choose the closest match:"\\\cmidrule{2-3}
& PE2 & Evaluate the given movies for genres, themes, narrative styles, tone, and characters to select the most similar one from the options.\\\midrule
\multirow{5}{*}{navigate}
& Iterative APE & Let's carefully assess each stage of our plan in sequence.\\\cmidrule{2-3}
& APO & Calculate the net distance from the starting point after following these step-by-step instructions, considering 'forward' as positive and 'backward' as negative movement.\\\cmidrule{2-3}
& PE2 & Let's calculate each movement's effect on our position. Start at zero and add or subtract steps as instructed, considering the direction each time to ensure accuracy.\\\midrule
\multirow{5}{*}{object\_counting}
& Iterative APE & Let's continue by taking systematic, sequential steps.\\\cmidrule{2-3}
& APO & Let's think step by step.\\\cmidrule{2-3}
& PE2 & Let's identify and count the instances of the specified category of items mentioned, tallying multiples, to determine their total quantity.\\\midrule
\multirow{5}{*}{penguins\_in\_a\_table}
& Iterative APE & Consider the procedure progressively.\\\cmidrule{2-3}
& APO & Let's think step by step.\\\cmidrule{2-3}
& PE2 & Let's think step by step.\\\midrule
\multirow{5}{*}{\begin{tabular}[c]{@{}l@{}}reasoning\_about\\\_colored\_objects\end{tabular}}
& Iterative APE & Consider each phase methodically.\\\cmidrule{2-3}
& APO & Let's think step by step.\\\cmidrule{2-3}
& PE2 & Let's think step by step and pay close attention to details such as colors and quantities.\\\midrule
\multirow{5}{*}{ruin\_names}
& Iterative APE & Let's proceed with our tasks one by one.\\\cmidrule{2-3}
& APO & Identify the funniest edit of the given name by selecting the option that best incorporates a pun or playful twist related to the original.\\\cmidrule{2-3}
& PE2 & Identify the most humorous edit by considering only puns or clever wordplay that creates a witty variation of the original name, excluding misspellings or simple pluralizations.\\\midrule
\multirow{5}{*}{\begin{tabular}[c]{@{}l@{}}salient\_translation\\\_error\_detection\end{tabular}
}
& Iterative APE & Let's take this gradually, step by step.\\\cmidrule{2-3}
& APO & Let's think step by step.\\\cmidrule{2-3}
& PE2 & Let's analyze the source and translation for errors. Check if there are any changes or inaccuracies in Named Entities, Numerical Values, Modifiers, Negation, Facts, or missing details. Identify which type of error occurs.\\\midrule
\multirow{5}{*}{snarks}
& Iterative APE & Evaluate each stage sequentially.\\\cmidrule{2-3}
& APO & Identify the sarcastic statement by considering the reversal of expectations and societal norms. Look for irony and implied meanings contrary to the literal words.\\\cmidrule{2-3}
& PE2 & To identify which statement is sarcastic, consider that sarcasm often means saying the opposite of what's true in a mocking way.\\\midrule
\multirow{5}{*}{sports\_understanding}
& Iterative APE & Let's think step by step.\\\cmidrule{2-3}
& APO & Determine if the sentence is plausible: Match the athlete's known sport, current activity status, and sport-specific terms to assess accuracy in context.\\\cmidrule{2-3}
& PE2 & Let's think step by step.\\\midrule
\multirow{5}{*}{temporal\_sequences}
& Iterative APE & Consider each stage methodically.\\\cmidrule{2-3}
& APO & Identify the time period when the person was not seen and the location was open. Exclude times when the person was observed elsewhere or the location was closed.\\\cmidrule{2-3}
& PE2 & Analyze the timeline to pinpoint time slots when the individual was not seen, which indicate when events could have occurred.\\\midrule
\multirow{5}{*}{\begin{tabular}[c]{@{}l@{}}tracking\_shuffled\\\_objects\_five\_objects\end{tabular}
}
& Iterative APE & We'll tackle this systematically, one stage at a time.\\\cmidrule{2-3}
& APO & Track ball swaps and position changes separately. List each swap, update positions and ball ownership after each, and determine final states for both.\\\cmidrule{2-3}
& PE2 & Let's carefully track each player's position swaps step by step to determine their final positions.\\\midrule
\multirow{5}{*}{\begin{tabular}[c]{@{}l@{}}tracking\_shuffled\\\_objects\_seven\_objects\end{tabular}
}& Iterative APE & Approach each stage with systematic thought.\\\cmidrule{2-3}
& APO & Let's think step by step.\\\cmidrule{2-3}
& PE2 & Let's carefully track each book exchange step by step to determine the final owner of each book.\\\midrule
\multirow{5}{*}{\begin{tabular}[c]{@{}l@{}}tracking\_shuffled\\\_objects\_three\_objects\end{tabular}
}& Iterative APE & Reflect on each stage individually.\\\cmidrule{2-3}
& APO & Let's think step by step.\\\cmidrule{2-3}
& PE2 & Let's analyze each position swap in sequence to determine the final positions. Confirm the last known positions of all players before concluding.\\\midrule
\multirow{5}{*}{web\_of\_lies}
& Iterative APE & Let's proceed with a methodical, stepwise approach.\\\cmidrule{2-3}
& APO & Given a sequence of people's statements about others' truthfulness, determine the truth status of the final person. Assume the first statement's truth is known. Apply logical negation for each liar's statement.\\\cmidrule{2-3}
& PE2 & To determine who is truthful, invert the claim of a liar and trust a truthful person's claim. Apply this logic until the last claim.\\\midrule
\multirow{5}{*}{word\_sorting}
& Iterative APE & Progress through each stage sequentially.\\\cmidrule{2-3}
& APO & Sort words alphabetically, ignoring case. Exclude 'List:' label. For same first letters, sort remaining letters. Output in lowercase.\\\cmidrule{2-3}
& PE2 & Let's analyze and sort words. Ignore words that are part of instructions, like "List:", and then arrange the remaining words in alphabetical order.\\

\end{longtable}

\begin{table*}[h]
\centering
\scalebox{0.7}{
\begin{tabular}{l|p{\textwidth}|c}
\toprule
Method & Final Prompt & Test Acc.\\\midrule
\rowcolor{gray!20}\multicolumn{3}{l}{Date Understanding (Generative)} \\\midrule
Zero-shot CoT & Let's think step by step. & 0.391 \\\midrule
Iterative APE & Let's dissect it and ponder over each phase. & 0.467 \\\midrule
APO & Determine the exact date from the scenario, considering cultural date formats, time zones, and periods. Use the provided date as a reference. Account for leading zeros, leap years, relative dates, and event-based time references. Provide the result in MM/DD/YYYY format. & 0.450 \\\midrule
PE2  & Analyzing the given information, let's calculate the solution. Remember to consider the context provided, such as references to 'today' or specific dates. & 0.544 \\\midrule
\rowcolor{gray!20}\multicolumn{3}{l}{Movie Recommendation (Multi-choice 2)} \\\midrule
Zero-shot CoT & Let's think step by step. & 0.570 \\\midrule
Iterative APE & Let's dissect it and consider every step in order. & 0.673 \\\midrule
APO & Identify the movie that shares the most significant themes and narrative structure with the given movies. Prioritize these factors over tone and pacing. Choose the most similar movie from the options, explaining your choice. & 0.750 \\\midrule
PE2 & Considering factors such as genre, director, actors, release period, audience target, animation style, and humor, analyze the similarities among the given movies and identify the movie from the options that shares the most similarities. & 0.790 \\\bottomrule
\end{tabular}
}
\caption{Results on Date Understanding and Movie Recommendation from BIG-bench Hard \citep{suzgun-etal-2023-challenging}. In these experiments, we use a format different from those in \citep{suzgun-etal-2023-challenging}. See \S\ref{app:analysis_format} for detailed discussion on the effect of task format. We use \texttt{gpt-3.5-turbo-instruct} as the task model and \texttt{gpt-4} as the prompt proposal model.}\label{table:bbh}
\end{table*}

%% file: tables/mistral_prompts.tex
\begin{table*}[h]
\centering
\scalebox{0.7}{
\begin{tabular}{l|p{\textwidth}|c}
\toprule
Method & Final Prompt & Test Acc.\\\midrule
\rowcolor{gray!20}\multicolumn{3}{l}{GSM8k} \\\midrule
Zero-shot CoT & Let's think step by step. & 0.481 \\\midrule
Iterative APE & We'll approach this methodically, proceeding one step at a time. & 0.497 \\\midrule
APO & Carefully analyze the details and perform precise arithmetic operations step by step, ensuring to apply the correct mathematical principles for accurate calculations. & 0.510\\\midrule
PE2  & Carefully calculate step by step, considering all details, including unique counts and overlaps. Accurately apply arithmetic to find the numerical answer, ensuring logical operations are correctly followed. & 0.505 \\\midrule
\rowcolor{gray!20}\multicolumn{3}{l}{MultiArith} \\\midrule
Zero-shot CoT & Let's think step by step. & 0.715 \\\midrule
Iterative APE & Proceed gradually, one step at a time. & 0.735 \\\midrule
APO & Calculate the answer using arithmetic. Round down where necessary. Correct any logical errors in reasoning. Provide the exact number. & 0.735\\\midrule
PE2  & Focus on accurately calculating totals and differences, considering factors like item conditions or groupings for precision. Round only if necessary, when dealing with practical fractions. & 0.743 \\\midrule
\rowcolor{gray!20}\multicolumn{3}{l}{BIG-bench Hard - Date Understanding} \\\midrule
Zero-shot CoT & Let's think step by step. & 0.36 \\\midrule
Iterative APE & Let's move forward by dividing it into manageable steps. & 0.48 \\\midrule
APO & Accurately calculate dates from given inputs, considering the current reference point (e.g., 'yesterday', 'today') and applying correct calendar arithmetic, including month lengths and leap years. Ignore irrelevant details and assumptions beyond provided information. & 0.48\\\midrule
PE2  & Let's think step by step. For date questions, calculate from the given date, then choose the closest matching option. Ignore irrelevant info. & 0.56 \\\midrule
\rowcolor{gray!20}\multicolumn{3}{l}{BIG-bench Hard - Hyperbaton} \\\midrule
Zero-shot CoT & Let's think step by step. & 0.52 \\\midrule
Iterative APE & Let's systematically go through every step. & 0.48 \\\midrule
APO & Classify adjective order: opinion, size, age, shape, color, origin, material, purpose. For words fitting multiple categories, prioritize purpose, then material. Ignore rare exceptions. & 0.72\\\midrule
PE2  & Choose the correct sentence using adjective order: opinion, size, age, shape, color, origin, material, purpose. Note: 'purpose' adjectives, like 'hiking', often come last. & 0.74 \\\midrule
\rowcolor{gray!20}\multicolumn{3}{l}{BIG-bench Hard - Temporal Sequences} \\\midrule
Zero-shot CoT & Let's think step by step. & 0.50 \\\midrule
Iterative APE & Let's progress by breaking it down into smaller, manageable parts. & 0.42 \\\midrule
APO & Identify when a person visited a place by excluding times they were seen elsewhere, considering their schedule, eyewitness sightings, and the place's hours. Only include unaccounted times. & 0.52\\\midrule
PE2  & Let's analyze the timeline and others' observations to deduce the only time slots not accounted for, indicating when the visit could have occurred. & 0.62 \\\midrule
\rowcolor{gray!20}\multicolumn{3}{l}{BIG-bench Hard - Word Sorting} \\\midrule
Zero-shot CoT & Let's think step by step. & 0.04 \\\midrule
Iterative APE & Let's tackle this systematically, advancing step by step. & 0.20 \\\midrule
APO & Alphabetically sort the words below, correcting any typos. Ignore capitalization, treat abbreviations and possessives normally. Exclude 'List:' from items. & 0.16\\\midrule
PE2  & Sort the given words alphabetically, but exclude 'List:' or similar formatting elements. Ensure every word is considered. & 0.28 \\\bottomrule
\end{tabular}
}
\caption{Results on six selected tasks. We use \texttt{Mistral-7B-Instruct-v0.2} as the task model and \texttt{gpt-4-turbo} as the prompt proposal model.}\label{table:mistral-prompts}
\end{table*}